\documentclass{article}

\PassOptionsToPackage{numbers,sort}{natbib}
\usepackage[main,final]{neurips_2026}

\usepackage{amsmath}
\usepackage{amssymb}
\usepackage{graphicx}
\usepackage{booktabs}
\usepackage{multirow}
\usepackage{colortbl}
\usepackage[table]{xcolor}
\usepackage{arydshln}
\usepackage[hypertexnames=false]{hyperref}
\usepackage{flafter}
\usepackage{float}
\usepackage[most]{tcolorbox}

\graphicspath{{assets/figures/}{assets/icons/}}

\definecolor{graybg}{RGB}{230, 230, 230} 
\definecolor{purplebg}{RGB}{235, 230, 250} 

\newtcolorbox{promptbox}[1]{
  enhanced,
  breakable,
  colback=white,
  colframe=black,
  colbacktitle=black,
  coltitle=white,
  fonttitle=\bfseries,
  boxrule=0.8pt,
  arc=2pt,
  left=8pt,
  right=8pt,
  top=6pt,
  bottom=6pt,
  toptitle=4pt,
  bottomtitle=4pt,
  title={#1},
  before skip=0.8em,
  after skip=0.8em
}

\title{MemCode: Discrete Semantic Representations for Long-Term Agent Memory}

\author{
Xin Li\textsuperscript{1} \quad Liang Hu\textsuperscript{1} \quad Duoqian Miao\textsuperscript{1}\quad Ming Peng\textsuperscript{1} \quad Qi Zhang\textsuperscript{1,2}\thanks{Corresponding Author}\\
  \textsuperscript{1}Tongji University \quad 
  \textsuperscript{2}WorldMind\\
  \texttt{\{lx\_cs, lianghu, dqmiao, pengming,~zhangqi\_cs\}@tongji.edu.cn} \\
}

\begin{document}
\maketitle
\raggedbottom

\begin{abstract}
Long-term memory remains a fundamental bottleneck for large language model (LLM) agents across sessions. Existing systems typically adopt a write-then-retrieve paradigm, storing context-dependent dialogue fragments and retrieving them via nearest-neighbor search in continuous embedding space. This design yields unstable access units and systematically misses logically related but lexically distant evidence. We propose MemCode, a long-term agent memory framework that couples retrieval-oriented writing with discrete associative retrieval. MemCode rewrites dialogue into self-contained atomic memory units (AMUs), maps AMUs into a learnable codebook with isomorphic semantic quantization (ISQ), and uses codebook co-occurrence topology as a retrieval path complementary to vector search. For long-term memory stores, MemCode separates online atomic writes from asynchronous offline consolidation. On LongMemEval-S and LoCoMo with GPT-4o-mini and Qwen3-30B-A3B-Instruct-2507 backbones, MemCode achieves the best overall accuracy while keeping per-dialogue memory construction within roughly 40$\sim$130 seconds, making it 3$\sim$21$\times$ faster than LightMem and over an order of magnitude faster than A-MEM and MemoryOS.

\end{abstract}

\section{Introduction}
\begin{figure}[t!]
\centering
\includegraphics[width=1\linewidth]{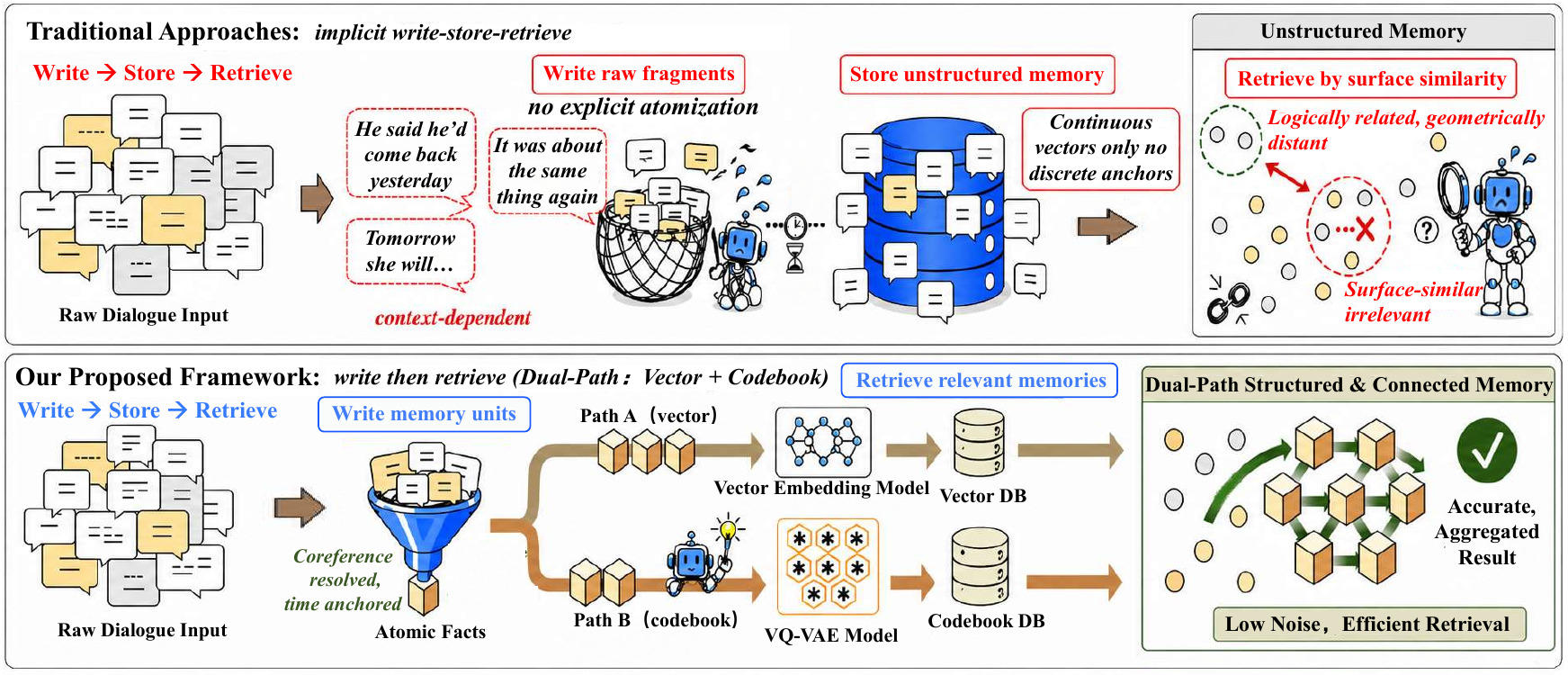}
\caption{Traditional memory pipelines store raw dialogue fragments and retrieve them primarily by surface similarity. Ours rewrites dialogue into atomic memory units and retrieves them through complementary vector and codebook paths.}
\vspace{-4mm}
\label{fig:motivation}
\end{figure}

Large language model (LLM) agents are rapidly evolving from single-turn systems into persistent assistants that interact with users over extended sessions~\citep{xi2023rise,wang2024agentsurvey}. However, neither model parameters nor context windows provide a satisfactory substrate for this need. Parameters are largely static after training, while long contexts incur quadratic attention costs~\citep{vaswani2017attention} and exhibit degraded utilization as sequences grow~\citep{liu2024lostmiddle}. It fundamentally elevates the role of agent memory that, distinct from both parameters and context, must accumulate, organize, and recall information across sessions to support personalization, consistency, and long-horizon reasoning.

As a result, external memory has emerged as a central component for agent systems. Most memory systems follow a retrieval-augmented generation (RAG) paradigm~\citep{lewis2020rag,izacard2021fid}: past interactions are written into storage, retrieved at query time, and re-injected into the model context. In this setting, two design choices are tightly coupled: 1) what to write, and 2) how to retrieve it. Prior systems address these choices primarily from memory persistence. Mem0~\citep{mem0_2025}, LangMem~\citep{langmem2025}, and A-MEM~\citep{amem_2025} improve writing and organization through Create, Read, Update, and Delete (CRUD) operations, typed memories, and evolvable note networks; MemoryOS~\citep{memoryos_2025} and LightMem~\citep{lightmem_2025} improve scalability through scheduling and compression. However, we argue that the key bottleneck for RAG-style long-term memory is not storage scale alone, but jointly modeling what to write and how to retrieve it, as illustrated in Figure~\ref{fig:motivation}.




This coupling becomes sharper as conversations grow. A memory item may be easy to store but expensive to use if it lacks the necessary context, such as missing speakers or times. Conversely, a compressed summary might answer broad questions but can erase the factual hooks needed for multi-hop recall. Our goal is to create independently retrievable memory units that maintain pointers to raw dialogue, avoiding irreversible abstraction bottlenecks.


Specifically, the \textbf{writing} pipeline is retrieval-oriented memory modeling, focusing on creating retrieval-accessible targets that remain understandable and indexable after separation from the original dialogue. Context-dependent fragments are weak storage units; instead, atomic facts are desirable. Long-term recall is usually cued by events and relations, not exact wording. In addition, the \textbf{retrieval} issue revolves around access representation. Most systems use continuous vectors for memory, retrieving content via top-$K$ nearest-neighbor searches~\citep{reimers2019sentencebert}, as illustrated by the upper pipeline in Figure~\ref{fig:motivation}. This works for close memories in surface semantics but inherits the geometric bias of continuous representations: logically related but lexically distant content may be missed, consistent with evidence that dense retrievers can be brittle when retrieval conditions shift across tasks~\citep{thakur2021beir}. In contrast, human memory distinguishes itself through pattern separation, allowing the hippocampus to encode similar continuous inputs into discrete traces for associative recall of partial cues~\citep{yassa2011pattern,tonegawa2015engram}. Systems that just mimic human multi-level memory without discrete anchors may still misalign architecture and retrieval.


Our proposed MemCode retains the RAG-style pipeline while adding a discrete representation layer alongside continuous vectors, utilized throughout writing, retrieval, and long-term maintenance. During writing, we employ LLM-based ingestion and abstraction to transform dialogue into self-contained AMUs with source pointers. Each AMU serves as a minimal semantic unit that can be independently retrieved. Independent semantic quantization (ISQ) anchors each AMU to a discrete symbolic state within a learnable codebook. During retrieval, global codebook co-occurrence offers an alternative access path for logically related yet geometrically distant memories, while source pointers enhance context reconstruction. MemCode also separates online writing from offline consolidation, allowing clustering and synthesizing AMUs without disrupting real-time retrieval.


\textbf{Contributions.}
Our contributions are fourfold. (1) We propose MemCode, a unified write--index--retrieve--consolidate framework that addresses the write--retrieve mismatch in long-term agent memory by organizing memory around atomic facts and discrete representations. (2) We introduce AMUs as the retrieval unit of this framework, turning raw dialogue into self-contained and source-grounded facts. 
(3) We pair AMUs with ISQ and codebook topology, giving each fact both a continuous vector interface and a discrete associative path for evidence beyond local embedding neighborhoods. (4) On LongMemEval-S, MemCode reaches 78.40\%/74.60\% Overall with GPT-4o-mini/Qwen3 after asynchronous consolidation; on LoCoMo it reaches 79.55\%/76.88\%, while keeping construction runtime roughly 3$\sim$21$\times$ faster than LightMem and over an order of magnitude faster than A-MEM and MemoryOS\@.
\section{Related Work}

\subsection{RAG-Based Long-Term Memory for Agents}
Most long-term memory schemes for agents are built on RAG, which combines external stores with generative models to reduce knowledge cutoffs and hallucination; in cross-session memory, the same pattern writes, indexes, and invokes historical interactions~\citep{karpukhin2020dpr}. \citet{gao2023rag} identify retrieval precision and context utilization as central limits that long-term memory inherits.

Prior work has explored diverse ways to write and organize memories. Mem0~\citep{mem0_2025} maintains a dynamic fact store through LLM-driven CRUD operations. A-MEM~\citep{amem_2025} organizes memories as an evolvable note network with explicit association edges, while LangMem~\citep{langmem2025} separates semantic, episodic, and procedural memory. These methods move beyond simple caches, but most machinery still sits above a continuous vector index. This leaves two gaps: 1) pronouns and relative-time expressions can make memory units hard to retrieve once detached from the original dialogue; 2) nearest-neighbor access is poorly matched to memories that are logically related but distant in surface semantics.
Other systems focus on storage management and lifecycle maintenance. MemoryOS~\citep{memoryos_2025} divides memory into short-, medium-, and long-term levels with heat-driven eviction, LightMem~\citep{lightmem_2025} lowers token cost through offline compression, and MEMORY-VQ~\citep{zemlyanskiy2024memoryvq} uses VQ-VAE to compress precomputed token representations for memory-augmented models. These methods improve storage use and throughput, but operate mainly at scheduling, compression, and eviction layers rather than retrieval-oriented memory organization. A long-term memory store can still grow redundant fragments with overlapping semantics, weakening later retrieval~\citep{packer2023memgpt}.

\subsection{VQ Codebooks as Discrete Memory Anchors}
In machine learning, vector-quantized variational autoencoders (VQ-VAE)~\citep{vqvae_2017} provide a standard discretization tool. VQ-VAE maps continuous encodings to discrete embeddings in a finite codebook through a nearest-neighbor quantizer and uses the straight-through estimator (STE) for the non-differentiable quantization step. A finite codebook encourages distinguishable semantic primitives and has been widely used in image generation and speech coding~\citep{zeghidour2021soundstream}. Discrete codebooks can therefore serve as compact anchors for high-dimensional continuous representations.
This codebook view is also compatible with neurocognitive memory intuition. \citet{yassa2011pattern} describe how the hippocampal dentate gyrus separates similar inputs into sparse traces, while the hippocampal CA3 subfield supports pattern completion from partial cues~\citep{tonegawa2015engram}. For agent memory, the relevant lesson is narrower: discrete anchors can separate similar facts and support associative retrieval beyond geometric nearest neighbors.
Dialogue memory, however, requires a retrieval-oriented use of VQ rather than direct lossy compression: because AMUs contain timestamps, entity identifiers, and state descriptions as factual anchors, MemCode applies stride-1 ISQ to preserve token-level alignment and uses the resulting discrete codes as associative retrieval keys.

\section{Methodology}
\begin{figure}[!t]
\centering
\includegraphics[width=0.99\linewidth]{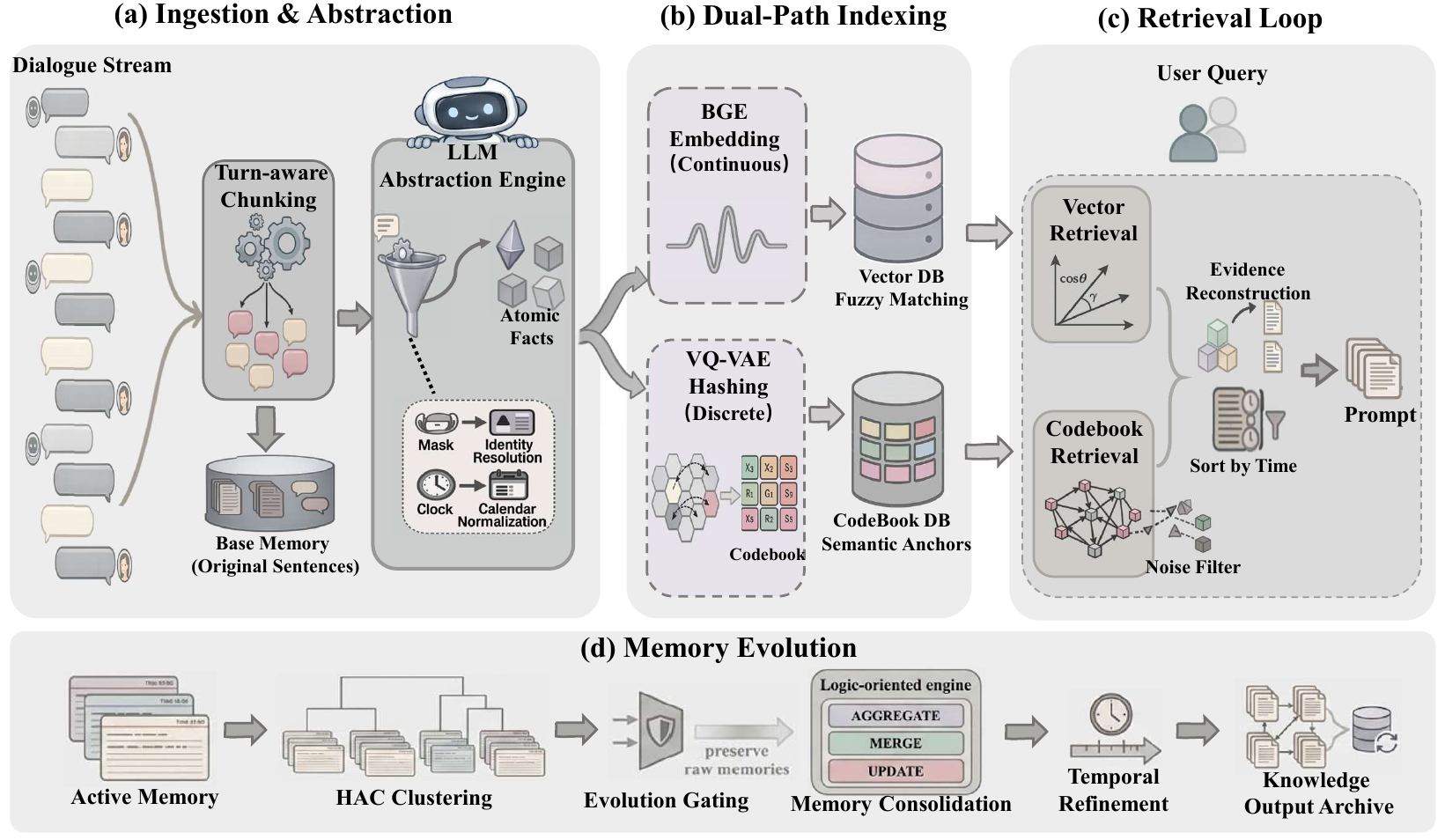}
\caption{MemCode architecture. Starting from raw dialogue streams, \textbf{Ingestion \& Abstraction} preserves source-grounded base memory and rewrites interactions into self-contained atomic memory units (AMUs). \textbf{Dual-Path Indexing} assigns each AMU both a continuous embedding and a discrete codebook fingerprint. During the \textbf{Retrieval Loop}, vector search provides high-precision matches, codebook topology adds associative evidence, and source pointers reconstruct richer context for answer generation. \textbf{Memory Evolution} then asynchronously consolidates long-term facts through clustering, deduplication, temporal refinement, and logic-oriented merging.}
\label{fig:architecture}
\end{figure}

\subsection{System Overview}
Figure~\ref{fig:architecture} summarizes MemCode. Ingestion canonicalizes dialogue into self-contained facts (Section~3.2). Dual-path indexing maps each fact to a dense vector and an ISQ codebook fingerprint (Section~3.3). Retrieval uses the dense vector for local matching, then uses codebook co-occurrence to add associated evidence and reconstruct source context (Section~3.4). Consolidation clusters, merges, and deduplicates memories offline (Section~3.5).
\subsection{Stream-Based Atomic Abstraction Framework}
\textbf{Interaction chunking \& context injection.}
The writing stage preserves streaming order. We represent the raw stream as $\mathcal{S}=\langle m_1,\ldots,m_N\rangle$, where $m_i=(r_i,t_i,c_i)$ contains role, timestamp, and text. Each compound message $c_i$ is split into atomic sentences $\mathbf{s}_i$~\citep{kiss2006sentence}, and messages are grouped into true interaction turns $\mathcal{B}=\{B_1,\ldots,B_{K_{\mathrm{turn}}}\}$. If $\mathcal{I}_k$ is the message-index set of turn $k$, then $B_k=\langle m_i:i\in\mathcal{I}_k\rangle$ in chronological order. Because strict turn segmentation can break coreference, MemCode processes the $k$-th turn with a suffix of $B_{k-1}$ as read-only background:
\begin{equation}
\mathcal{H}_k=\text{Suffix}(B_{k-1}, \omega), \qquad
\mathcal{X}_k = \underbrace{\mathcal{H}_k}_{\text{Background Context}}
\oplus \underbrace{B_k}_{\text{Active Target}} .
\end{equation}
Here, $\mathcal{H}_k$ contains the last $\omega$ sentences from the previous turn, with $\mathcal{H}_1=\emptyset$ for the first turn; $\omega$ is the extraction-background budget, and $\oplus$ denotes concatenation. The background resolves local coreference but is excluded from fact extraction.

\textbf{Semantic canonicalization \& atomic extraction.}
Given $\mathcal{X}_k$, the abstraction module rewrites the active turn $B_k$ into self-contained facts $\mathcal{F}_k$ that remain retrievable after being detached from the dialogue. We model this as a generative mapping conditioned on the read-only background and metadata extracted from $\mathcal{X}_k$:
\begin{equation}
\mathcal{F}_k = \Phi(B_k \mid \mathcal{H}_k, T_{anchor}, \mathcal{R}_{roles})
\end{equation}
Here, $\Phi$ is the LLM abstraction engine. $T_{anchor}$ is derived from message timestamps in $\mathcal{X}_k$ and anchors relative time, while $\mathcal{R}_{roles}$ is the role--speaker map extracted from message roles and used with background to resolve pronouns~\citep{lee2017e2ecoreference}. Each output AMU $u\in\mathcal{F}_k$ is denoted as:
\begin{equation}
u = (c_{fact},\ t_{abs},\ \Lambda_{src})
\end{equation}
where $c_{fact}$ is the canonicalized fact generated from $B_k$, $t_{abs}$ is the absolute timestamp obtained by resolving relative time against $T_{anchor}$, and $\Lambda_{src}$ points to original spans in $B_k$ and, when needed for grounding, $\mathcal{H}_k$. During retrieval, these pointers recover nearby utterances when an atomic fact matches the query but is too terse for answer generation.
\subsection{Fine-Grained Discrete Semantic Indexing}
\textbf{Isomorphic semantic quantization.}
To build an associative discrete index, MemCode follows VQ-VAE and treats the codebook as a latent state basis~\citep{defossez2022encodec}. For AMU text $x=c_{fact}$, ISQ obtains encoder states $H=(h_1,\ldots,h_L)$ from a T5-base encoder fine-tuned on DailyDialog~\citep{li2017dailydialog}. Here, stride denotes the projection step over token positions. A learned stride-1 projection maps each state to $z_t=W_p h_t+b_p$, giving $Z=(z_1,\ldots,z_L)$ and preserving $L_{\mathrm{out}}=L$ so timestamps, numbers, and entity tokens are not dropped.
We then introduce a learnable Euclidean codebook $\mathcal{E}=\{e_j\}_{j=1}^{N_c}$, where $e_j$ is the $j$-th code vector and $N_c$ is the codebook size. Each projected vector $z_t$ is assigned to its nearest code:
\begin{equation}
    \kappa_t = \operatorname*{argmin}_{j \in \{1,\ldots,N_c\}}
    \|z_t-e_j\|_2,
    \qquad
    \bar{z}_t = e_{\kappa_t}.
\end{equation}
Here, $\kappa_t$ is the discrete code id and $\bar{z}_t$ is the quantized vector used in place of $z_t$ by the decoder. Backpropagation uses the straight-through estimator (STE)~\citep{bengio2013ste}.

\textbf{Conversational codebook learning \& semantic fingerprinting.}
The codebook is trained with predictive coding rather than input reconstruction: for each context--response pair, the model predicts the response from the quantized context representation~\citep{oord2018cpc}, preserving reasoning cues while filtering shallow lexical noise. The loss is:
\begin{equation}
    \mathcal{L}_{total} =
    -\sum_{r=1}^{|y|}\log p_\theta(y_r\mid y_{<r},\bar{Z})
    + \lambda_{\mathrm{vq}}\sum_{t=1}^{L}\ell_{\mathrm{vq}}(z_t,\bar{z}_t).
\end{equation}
Here, $y$ is the target response, $\theta$ denotes the decoder parameters, $\bar{Z}=(\bar{z}_1,\ldots,\bar{z}_L)$ is the quantized context sequence, $\ell_{\mathrm{vq}}=\|\operatorname{sg}[z_t]-\bar{z}_t\|_2^2+\beta\|z_t-\operatorname{sg}[\bar{z}_t]\|_2^2$, $\operatorname{sg}[\cdot]$ stops gradients, $\beta$ is the commitment weight, and $\lambda_{\mathrm{vq}}$ balances generation and quantization. The codebook $\mathcal{E}$ is updated by EMA~\citep{razavi2019vqvae2}; low-frequency codewords are reinitialized from current-batch encoder outputs.
At inference time, adjacent repeats in $\kappa_{1:L}$ are removed to form a compact fingerprint:
\begin{equation}
    S_{fp}(u)=
    \left\langle \kappa_t \;:\; 1\leq t\leq L,\ 
    t=1\ \text{or}\ \kappa_t\neq\kappa_{t-1}\right\rangle .
\end{equation}
The sequence $S_{fp}(u)$ is the discrete fingerprint of AMU $u$. Its unique-code set $U_{fp}(u)=\{a:a\in S_{fp}(u)\}$ is used for topology construction, so repeated code ids are counted once per AMU when building co-occurrence statistics.
Quantization does not replace the stored AMU text: the original $c_{fact}$ remains the evidence presented to the answer generator, while $S_{fp}(u)$ serves only as an auxiliary retrieval key.
\subsection{Topology-Aware Dual-Path Retrieval}
\textbf{Implicit topology \& associative inference.}
The discrete fingerprints define an implicit topology through codebook co-occurrence, complementing the geometric locality of dense vectors. Let $\mathcal{F}=\bigcup_{k=1}^{K_{\mathrm{turn}}}\mathcal{F}_k$ be the global AMU store and $N_u=|\mathcal{F}|$. We keep a global matrix $\mathcal{M}$ over code ids:
\begin{equation}
    \mathcal{M}_{ab} =
    \sum_{u\in\mathcal{F}}
    \mathbf{1}[a\in U_{fp}(u)]\,\mathbf{1}[b\in U_{fp}(u)]\,\mathbf{1}[a\neq b],
    \qquad
    n(a)=\sum_{u\in\mathcal{F}}\mathbf{1}[a\in U_{fp}(u)] .
\end{equation}
Thus $\mathcal{M}_{ab}$ counts how often distinct ids $a$ and $b$ co-occur in the same AMU fingerprint, while $n(a)$ counts how many AMUs contain source id $a$; diagonal self-transitions are masked~\citep{levy2014pmi}. The one-hop transition probability is:
\begin{equation}
    P(b \mid a) = \frac{\mathcal{M}_{ab}}{\sum_{b'\neq a} \mathcal{M}_{ab'} + \epsilon}.
\end{equation}
Here, $\epsilon$ is a small smoothing constant for codes with no outgoing co-occurrence. During inference, the query is projected and deduplicated by the same ISQ module, giving query code set $Q_{fp}$. Treating $P(b \mid a)$ as the entry $P_{ab}$ of the code-to-code transition matrix, MemCode scores activation from a source id $a\in Q_{fp}$ to an associated target id $b$ across up to $H_r$ topology hops:
\begin{equation}
    S_{\mathrm{code}}(b \mid a) =
    \underbrace{\log \left( \frac{N_{u}+1}{n(a) + 1} \right)}_{\text{source IDF}}
    \cdot
    \max_{1\leq h\leq H_r}
    \underbrace{[P^h]_{ab}}_{\text{$h$-hop transition}}
    \cdot\ \alpha^h
\end{equation} 
where $\alpha\in(0,1)$ controls decay during neighbor expansion and $H_r$ is the fixed retrieval expansion depth. We use $H_r=1$ in all experiments, preventing recursive topology propagation from amplifying noisy associations. The source-side IDF downweights frequent query anchors~\citep{robertson2009bm25}, while the transition term activates codes reached by repeated co-occurrence.

\textbf{Hybrid evidence retrieval \& context reconstruction.}
The final retrieval stage merges the two evidence paths through structured supplementation. The dense path (Path A) retrieves $K_v$ atomic facts by nearest-neighbor search over their dense vectors, forming $\mathcal{R}_{vec}$. The codebook path (Path B) scores each AMU by the best code in its fingerprint:
\begin{equation}
    S_{\mathrm{amu}}(u)=
    \max_{\substack{a\in Q_{fp}\\ b\in U_{fp}(u)}}
    S_{\mathrm{code}}(b\mid a).
\end{equation}
It then selects $K_c$ supplementary facts from $\mathcal{F}$, excluding Path A memories:
\begin{align}
    &\mathcal{R}_{topo}
    = \operatorname{Top\text{-}K}_{c}\!\left(
        \mathcal{F} \setminus \mathcal{R}_{vec}
        \;;\; \text{ranked by } S_{\mathrm{amu}}(u) \downarrow
    \right), \notag \\
    &\mathcal{R}_{hybrid}
    = \mathcal{R}_{vec} \cup \mathcal{R}_{topo}.
\end{align}
Here, $K_v$ and $K_c$ are the fixed dense-path and codebook-path evidence budgets, and $\downarrow$ denotes descending order by score. Greedy topology selection adds evidence associated in codebook space but absent from dense candidates. MemCode then follows $\Lambda_{src}$ back to original spans and orders them chronologically.
The merge is intentionally asymmetric: Path A provides high-precision neighbors, while Path B only supplements it. This keeps popular associations from overwhelming the answer context and makes the ablation interpretable under a fixed $K_v{+}K_c$ evidence budget.
In Appendix, Figures~\ref{fig:appendix_codebook_usage}--\ref{fig:appendix_codebook_topology} further show learned codes form a concentrated but non-collapsed topology, supporting asymmetric use of Path B.
\subsection{Asynchronous Memory Consolidation}
Online atomic writes can accumulate duplicate facts, outdated states, and retrieval interference. MemCode handles this with an asynchronous consolidation engine outside the online path. It backfills dense vectors, filters invalid entries, and applies hierarchical agglomerative clustering (HAC) based on cosine distance~\citep{ward1963hac}:
\begin{equation}
\mathcal{G}=\operatorname{HAC}\!\left(\mathcal{F};\ d(u_i,u_j)=1-\frac{\mathbf{v}_i\cdot\mathbf{v}_j}{\|\mathbf{v}_i\|\|\mathbf{v}_j\|},\ d_{\max}=1-\tau\right).
\end{equation}
Here, $\mathbf{v}_i$ is the dense vector of AMU $u_i$ produced by the same encoder used by Path A, $\tau$ is the configured minimum cosine similarity for grouping, and $\mathcal{G}$ is the candidate cluster set. Large clusters are split by time. Before entering the LLM, a gate skips count-sensitive clusters, restricts update-sensitive clusters, and avoids cross-speaker mixing. Passing clusters are assigned to AGGREGATE, MERGE, or UPDATE, with fewer new facts than archived facts.
After write-back, Pointer Forwarding marks old facts as archived and records their new fact indices, forming a two-level evidence chain with the original $\Lambda_{src}$. At retrieval, old facts are replaced by new ones. If the temporal span of a cluster $g \in \mathcal{G}$ exceeds the minute level, the new fact is prefixed with \texttt{[Time Span: $T_{start}$ to $T_{end}$]}, where $T_{start} = \min_{u \in g} t_{abs}(u)$ and $T_{end} = \max_{u \in g} t_{abs}(u)$. The archive policy combines consolidation mode, number of parent facts, time span, and speaker source to decide whether to replace or retain a fact, reducing the risk of irreversible information loss.

\section{Experiments}
\begin{table}[!t]
\centering
\small
\caption{Accuracy and memory-construction cost on LongMemEval-S. We report Overall/category ACC ($\uparrow$), runtime per dialogue ($\downarrow$), Total Tokens, and Calls. \textit{Asyn.} denotes asynchronous consolidation. Bold and underline denote the best and 2$^{nd}$-best results within each backbone.}
\label{tab:main_results}
\resizebox{\linewidth}{!}{%
\begin{tabular}{lcccccccccc}
\toprule
\multirow{2}{*}{\textbf{Method}} & \textbf{Runtime} $\downarrow$  & \textbf{Total Tokens} $\downarrow$ & \textbf{Calls} $\downarrow$ & \textbf{Overall} $\uparrow$ & \textbf{Temporal} & \textbf{Multi-Session} & \textbf{Know.-Upd.} & \textbf{Single-User} & \textbf{Single-Asst.} & \textbf{Single-Pref.} \\
& \textbf{(s)} & \textbf{(k)} &  & \textbf{ACC (\%)} & ($n=133$) & ($n=133$) & ($n=78$) & ($n=70$) & ($n=56$) & ($n=30$) \\
\midrule
\rowcolor{graybg} \multicolumn{11}{c}{\includegraphics[height=1.2em]{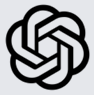} \textbf{ GPT-4o-mini}} \\
\midrule
\textit{Full Text}    & - & - & - & 56.80 & 31.58 & 45.45 & 76.92 & 87.14 & 89.29 & 36.67 \\
\textit{Naive RAG}    & - & - & - & 61.00 & 39.85 & 48.48 & 67.95 & 90.00 & \textbf{98.21} & 53.33 \\
\textit{LangMem}      & 2293.70 & 1102.16 & 520.62 & 37.20 & 15.79 & 20.30 & 66.67 & 60.00 & 46.43 & 60.00 \\
\textit{A-MEM}        & 5132.06 & 1605.81 & 986.55 & 62.60 & 47.36 & 48.87 & 64.11 & \underline{92.86} & \underline{96.43} & 46.67 \\
\textit{MemoryOS}     & 8030.04 & 2991.75 & 2938.41 & 44.80 & 32.33 & 31.06 & 48.72 & 80.00 & 64.29 & 30.00 \\
\textit{Mem0}         & 4248.49 & 1152.62 & 811.57 & 53.61 & 40.15 & 46.21 & 70.12 & 81.43 & 41.07 & 60.00 \\
\textit{LightMem}     & 283.76 & \textbf{28.25} & \textbf{18.43} & 68.64 & 67.18 & \textbf{71.74} & \underline{83.12} & 87.14 & 32.14 & 68.18 \\
\midrule
\textit{MemCode}         & \textbf{74.05} & \underline{318.41} & \underline{248.84} & \underline{77.20} & \textbf{69.92} & 65.41 & 82.05 & \underline{92.86} & 92.86 & \underline{83.33} \\
\textit{MemCode + Asyn.} & \underline{82.98} & 328.05 & 274.37 & \textbf{78.40} & \underline{67.67} & \underline{69.17} & \textbf{83.33} & \textbf{95.71} & 91.07 & \textbf{90.00} \\
\midrule
\rowcolor{purplebg} \multicolumn{11}{c}{\includegraphics[height=1.2em]{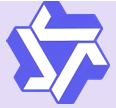} \textbf{ Qwen3-30B-A3B-Instruct-2507}} \\
\midrule
\textit{Full Text}    & - & - & - & 54.80 & 33.08 & 35.61 & 76.92 & 82.86 & 87.50 & 50.00 \\
\textit{Naive RAG}    & - & - & - & 60.80 & 36.84 & 47.73 & 65.38 & \underline{91.43} & \textbf{98.21} & 70.00 \\
\textit{LangMem}      & 3237.16 & 1430.02 & 495.12 & 50.80 & 37.60 & 38.35 & 67.95 & 78.57 & 42.86 & 70.00 \\
\textit{A-MEM}        & 5367.51 & 1864.93 & 989.30 & 65.20 & 51.88 & 51.12 & \underline{76.93} & 90.00 & \underline{96.43} & 40.00 \\
\textit{MemoryOS}     & 8721.78 & 2944.97 & 2922.28 & 49.60 & 28.57 & 36.84 & 61.54 & 72.86 & 92.86 & 33.33 \\
\textit{Mem0}         & 2239.94 & 1001.90 & 722.76 & 39.51 & 41.94 & 28.13 & 28.57 & 55.32 & 26.09 & \underline{81.82} \\
\textit{LightMem}     & 357.13 & \textbf{25.15} & \textbf{16.54} & 61.95 & 54.20 & 51.91 & 66.67 & 80.00 & 31.25 & 80.00 \\
\midrule
\textit{MemCode}         & \textbf{78.05} & \underline{356.41} & \underline{248.84} & \underline{73.80} & \underline{57.89} & \underline{64.66} & \textbf{78.21} & \textbf{94.29} & 92.86 & \textbf{90.00} \\
\textit{MemCode + Asyn.} & \underline{95.02} & 363.59 & 265.94 & \textbf{74.60} & \textbf{60.90} & \textbf{66.17} & \textbf{78.21} & \underline{91.43} & 92.86 & \textbf{90.00} \\
\bottomrule
\end{tabular}%
}
\vspace{-4mm}
\end{table}
\subsection{Experimental Setting}
\textbf{Datasets and protocol}. We evaluate long-term agent memory on LoCoMo~\citep{maharana2024locomo} and LongMemEval-S~\citep{wu2024longmemeval}. LongMemEval-S stresses dynamic multi-session retrieval and updating, while LoCoMo covers multi-hop, temporal, open-domain, and single-hop questions. We use an incremental protocol that mimics deployment: the system ingests conversations turn by turn in chronological order and cannot access the full history in advance.
\textbf{Baselines and implementation}. Baselines include Full Text, Naive RAG, LangMem, A-MEM, MemoryOS, Mem0, and LightMem; LangMem is reported on LongMemEval-S and omitted from the LoCoMo table. Full Text provides the controlled full-history self-attention alternative under the same answer backbone. All methods use GPT-4o-mini~\citep{openai2024gpt4omini} and Qwen3-30B-A3B-Instruct-2507~\citep{qwen3_2025}\@. MemCode's discrete representation uses a VQ-VAE initialized from T5-base~\citep{raffel2020t5} and trained on DailyDialog with a single 32GB GPU; its continuous representation uses 768-dimensional bge-base-en-v1.5 embeddings~\citep{bge2024}\@. We use only the official DailyDialog training split (11,118 dialogues; 64,935 context--response pairs) to learn reusable local conversational patterns. The codebook is frozen before memory evaluation; no benchmark histories, questions, answers, labels, or retrieval traces enter its optimization. Online ingestion and consolidation are constrained to a consumer RTX 3050 Ti GPU\@.
\textbf{Metrics}. For task performance, we follow the common LLM-as-a-Judge protocol~\citep{zheng2023judging}: GPT-4o-mini compares each output with the reference answer or key topic points, and we report overall and category accuracy. For efficiency, we report average runtime, Calls, and Total Tokens for memory construction only; after retrieval, all methods use a single LLM generation call, while vector retrieval and codebook inverted lookup are computationally negligible and depend little on external API conditions. In long-term agent settings, the bottleneck is therefore the construction of structured historical memory rather than millisecond-level differences in a single query.
Appendix Tables~\ref{tab:appendix_compute}--\ref{tab:appendix_implementation_budget} report the compute environments, offline VQ-VAE training settings, and dataset-specific retrieval budgets; Appendix~\ref{sec:appendix_prompts} reports the prompts.

\subsection{Main Results}
Tables~\ref{tab:main_results} and~\ref{tab:locomo_results} show that MemCode improves the accuracy--efficiency trade-off across both backbones and benchmarks. It achieves the highest overall accuracy while reducing construction runtime to roughly $1/21$--$1/3$ of LightMem's runtime and 1--2 orders of magnitude below A-MEM and MemoryOS\@. The category breakdown shows where the gains arise: atomic writing improves precise localization and speaker-aware facts, the codebook path supports structured temporal and compositional retrieval, and consolidation partially restores long-range continuity lost by atomic segmentation.
Appendix Tables~\ref{tab:appendix_backbone_accuracy}--\ref{tab:appendix_backbone_pairwise} further show that backbone choice changes the memory store itself: Table~\ref{tab:appendix_backbone_memory_density} shows that Qwen3 writes denser AMU/codebook traces, while Tables~\ref{tab:appendix_backbone_accuracy} and~\ref{tab:appendix_backbone_pairwise} connect this density to better open-ended coverage but more distractors for exact temporal and multi-hop evidence.
\begin{table}[!t]
\centering
\small
\renewcommand{\arraystretch}{0.86}
\caption{Category-level accuracy and memory-construction cost on LoCoMo. \textit{Asyn.} denotes asynchronous consolidation. \textbf{Bold} and \underline{underline} denote the best and 2$^{nd}$-best results on each backbone.}
\label{tab:locomo_results}
\resizebox{0.94\linewidth}{!}{%
\begin{tabular}{lcccccccc}
\toprule
\multirow{2}{*}{\textbf{Method}} & \textbf{Runtime} $\downarrow$ & \textbf{Total Tokens} $\downarrow$ & \textbf{Calls} $\downarrow$ & \textbf{Overall} $\uparrow$ & \textbf{Multi} & \textbf{Open} & \textbf{Single} & \textbf{Temp} \\
& \textbf{(s)} & \textbf{(k)} &  & \textbf{(ACC \%)} & ($n=282$) & ($n=96$) & ($n=841$) & ($n=321$) \\
\midrule
\rowcolor{graybg} \multicolumn{9}{c}{\includegraphics[height=1.2em]{icon_gpt4o_mini.png} \textbf{ GPT-4o-mini}} \\
\midrule
\textit{Full Text}                 & - & - & - & 73.83 & \underline{68.79} & \textbf{56.25} & 86.56 & 50.16 \\
\textit{Naive RAG}                 & - & - & - & 63.64 & 55.32 & 47.92 & 70.99 & 56.39 \\
\textit{A-MEM}                    & 6060.73 & 1149.43 & 1175.47 & 64.16 & 56.03 & 31.25 & 72.06 & 60.44 \\
\textit{MemoryOS (eval)}          & 2422.05 & 287.00 & 553.45 & 58.25 & 56.74 & 45.83 & 67.06 & 40.19 \\
\textit{MemoryOS (pypi)}          & 3332.59 & 526.48 & 1016.06 & 54.87 & 52.13 & 43.75 & 63.97 & 36.76 \\
\textit{Mem0}                     & 4432.87 & 1693.39 & 1602.20 & 61.69 & 56.38 & 43.75 & 66.47 & 59.19 \\
\textit{LightMem} & 848.49 & \textbf{99.76} & \textbf{41.65} & 71.95 & 62.41 & 44.79 & 77.41 & 74.14 \\
\midrule
\textit{MemCode}                     & \textbf{40.70} & \underline{189.32} & \underline{287.10} & \underline{78.12} & 64.89 & 43.75 & \underline{87.04} & \textbf{76.64} \\
\textit{MemCode + Asyn.}         & \underline{109.40} & 266.59 & 439.60 & \textbf{79.55} & \textbf{69.15} & \underline{51.04} & \textbf{87.87} & \underline{75.39} \\
\midrule
\rowcolor{purplebg} \multicolumn{9}{c}{\includegraphics[height=1.2em]{icon_qwen3_30b_a3b.png} \textbf{ Qwen3-30B-A3B-Instruct-2507}} \\
\midrule
\textit{Full Text}                 & - & - & - & 74.87 & \underline{69.86} & \underline{57.29} & 87.40 & 51.71 \\
\textit{Naive RAG}                 & - & - & - & 66.95 & 62.41 & \underline{57.29} & 76.81 & 47.98 \\
\textit{A-MEM}                    & 5543.90 & 1626.80 & 1175.40 & 56.10 & 57.45 & 43.75 & 67.90 & 27.73 \\
\textit{MemoryOS (eval)}          & 1269.70 & 361.51 & 414.70 & 61.04 & 62.77 & 51.04 & 72.29 & 33.02 \\
\textit{MemoryOS (pypi)}          & 1982.20 & 666.35 & 1004.60 & 51.30 & 52.48 & 40.62 & 61.59 & 26.48 \\
\textit{Mem0}                     & 4540.70 & 1799.40 & 1614.50 & 43.31 & 42.91 & 46.88 & 46.37 & 34.58 \\
\textit{LightMem} & 815.70 & \textbf{99.87} & \textbf{29.10} & 71.36 & \textbf{70.57} & \textbf{60.42} & 79.19 & 54.83 \\
\midrule
\textit{MemCode}                     & \textbf{41.14} & \underline{206.10} & \underline{287.10} & \underline{76.36} & 59.22 & 52.08 & \underline{88.11} & \textbf{67.91} \\
\textit{MemCode + Asyn.}         & \underline{129.69} & 339.81 & 517.50 & \textbf{76.88} & 66.31 & 48.96 & \textbf{88.59} & \underline{63.86} \\
\bottomrule
\end{tabular}
}
\vspace{-4mm}
\end{table}
\textbf{LongMemEval-S}. This benchmark contains roughly 30--40 sessions and about 115K tokens, with an emphasis on retrieval and updating in dynamic memory. Under GPT-4o-mini, MemCode reaches 77.20\% Overall, 8.56 points above LightMem; asynchronous consolidation raises it to 78.40\%. The same trend appears under Qwen3 (73.80\% $\to$ 74.60\%). Category results clarify the source of the gains. Absolute timestamps turn some temporal questions into constrained single-hop retrieval, which helps Temporal and Single-User. LightMem remains stronger on Multi-Session (71.74\% vs.\ 65.41\%), where paragraph compression preserves cross-session continuity; AGGREGATE partially recovers it to 69.17\%. MemCode is much stronger on Single-Assistant (92.86\% vs.\ 32.14\%), indicating the value of speaker binding.

In efficiency, MemCode takes about 74 seconds per dialogue and about 83 seconds with consolidation. Its Calls and Tokens exceed LightMem's, but the runtime is lower because atomic extraction is parallel, whereas batch compression is sequential. Tokens are closer to API cost; Runtime is closer to deployment latency.
\textbf{LoCoMo}. LoCoMo averages 304.9 dialogue turns and 9209.2 tokens. MemCode reaches 78.12\%/76.36\% Overall under GPT-4o-mini/Qwen3, and asynchronous consolidation improves this to 79.55\%/76.88\%. Single tasks improve over LightMem (87.04\% vs.\ 77.41\%), supporting atomic facts and spatiotemporal resolution in single-hop localization. On \textit{Multi} tasks, codebook topology under GPT-4o-mini exceeds LightMem and Naive RAG (64.89\% vs.\ 62.41\%/55.32\%), consistent with discrete retrieval recovering implicit cross-segment dependencies~\citep{trivedi2022musique}. Full Text and LightMem remain competitive on \textit{Multi} as a complete or compressed narrative context preserves logical connections; retrieval methods remain limited by recall completeness.

Open tasks expose a boundary of abstraction: mental states, implicit emotions, and motivations are difficult to fully propositionalize, so Full Text retains an advantage. Consolidation raises Open to 51.04\%, but the synthesized material still comes from extracted facts, leaving limited headroom. In efficiency, MemCode needs only 40.70 seconds on LoCoMo, about one twentieth of LightMem's runtime; with consolidation, 109.40 seconds remains below all baselines, so the offline stage retains the same parallelism advantage.
\textbf{Boundary of consolidation}. Consolidation synthesizes fragmented facts into denser memories, improving LoCoMo Multi by 4.26 and 7.09 points. UPDATE also reduces interference by archiving outdated states. The trade-off is temporal granularity: multiple time points may be compressed into one entry, causing a slight Temporal drop. Since extraction already normalizes absolute time, temporal results after consolidation still outperform or match most baselines. Qwen3 shows larger Multi gains but also a clearer Temporal drop, indicating that different backbones use compressed memories differently. This frames consolidation as an exchange between denser global memory and finer temporal resolution, not a uniformly monotonic gain.
Appendix Table~\ref{tab:appendix_repeated_run_variability} shows that the LoCoMo GPT-4o-mini result is stable across repeated runs.

Taken together, the two benchmarks separate two regimes. When answers are grounded in explicit entities, dates, speaker bindings, or linked facts, MemCode's AMUs and discrete topology outperform raw retrieval and compressed-memory baselines. When answers depend on implicit affect or broad narrative continuity, Full Text and LightMem remain competitive; consolidation narrows this gap by restoring continuity at a small temporal-granularity cost. The ablation below tests the retrieval path directly, while the appendix adds repeated-run stability (Table~\ref{tab:appendix_repeated_run_variability}), dense-embedding sensitivity (Table~\ref{tab:appendix_embedding_sensitivity}), retrieval novelty (Figure~\ref{fig:case_study}), and codebook-space structure (Figures~\ref{fig:appendix_codebook_usage}--\ref{fig:appendix_codebook_topology}).
\subsection{Ablation Study}
\begin{figure}[!tbp]
\centering
  \includegraphics[width=0.94\linewidth]{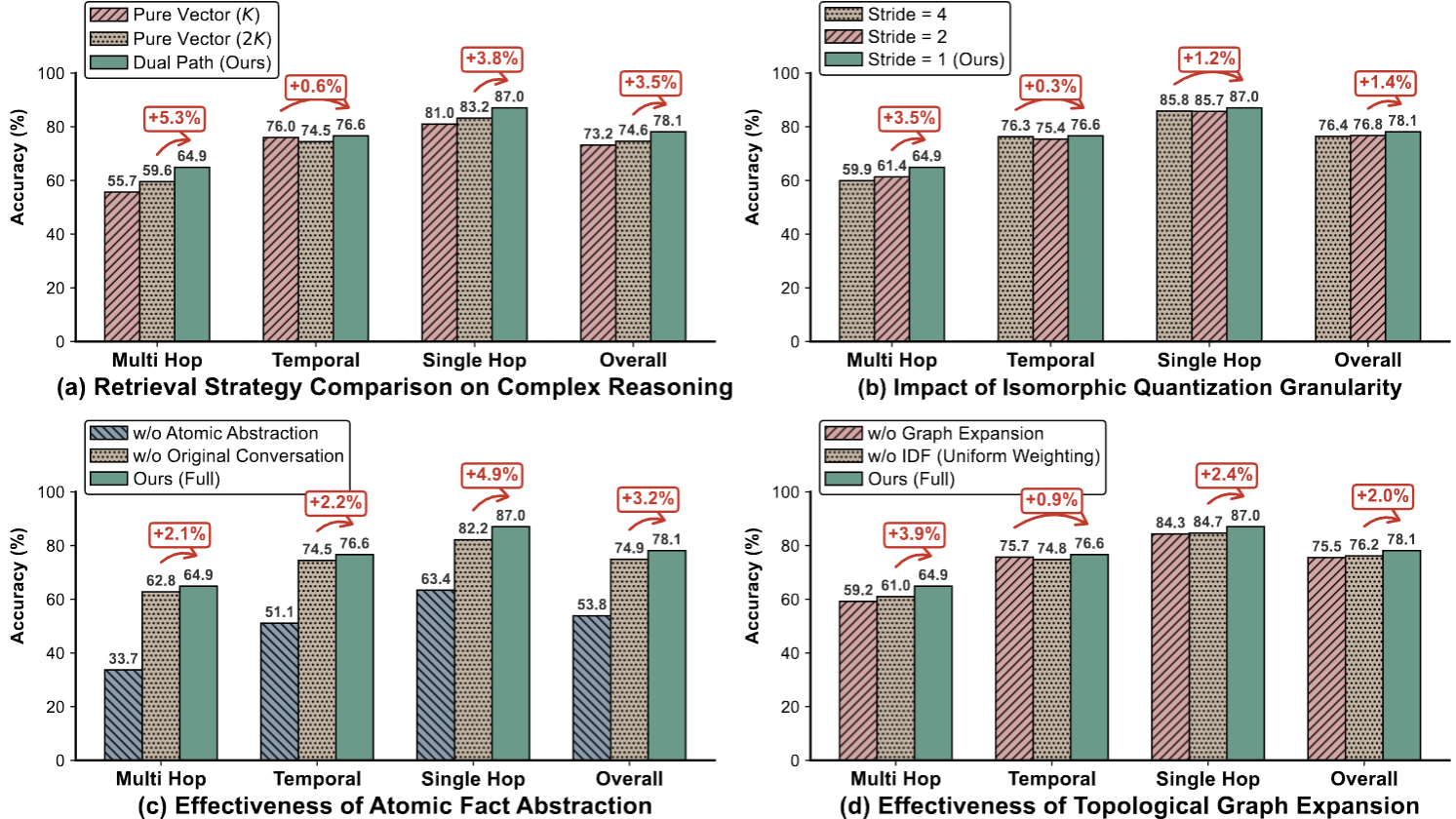}
  \caption{\label{fig:ablation}Ablation on LoCoMo (GPT-4o-mini): retrieval path, stride, ingestion/context reconstruction, and topology/IDF weighting.}
\end{figure}

We run ablations on LoCoMo with GPT-4o-mini. Figure~\ref{fig:ablation}(a) shows that doubling vector Top-$K$ adds only 1.4\%, whereas dual-path retrieval improves Multi-Hop by 5.3 points. The limiting factor is therefore not candidate count, but continuous neighborhoods missing logical associations in the discrete topology. Figure~\ref{fig:ablation}(b) shows that coarser stride hurts Temporal and Multi-Hop while barely affecting Single-Hop, confirming that timestamps and entity IDs are core memory tokens rather than implementation details.

Figure~\ref{fig:ablation}(c) shows that removing atomic-fact abstraction drops Overall to 53.8\% and Multi-Hop by 31.2\%, indicating that context-dependent writes are weak retrieved evidence. Removing raw context costs only 3.2\%, meaning that AMUs carry the main retrievable semantics while source reconstruction mainly improves evidence completeness. This confirms that the quality of the write unit is the primary driver of retrieval success, and highlights the importance of precise temporal and speaker grounding. Figure~\ref{fig:ablation}(d) shows that topology expansion provides long-range activation, while IDF suppresses high-frequency anchors; removing either component reduces Multi-Hop, and the full model benefits from combining recall expansion with noise control. Together, these results illustrate how structured retrieval paths complement dense embeddings to handle compositional and temporally distributed evidence.

\begin{table}[!tbp]
\centering
\footnotesize
\setlength{\tabcolsep}{8pt}
\caption{Component attribution on LoCoMo Multi-Hop (GPT-4o-mini, $n=282$). All settings use the same base AMU store and a total evidence budget of 40. Counts of correct answers are shown in parentheses.}
\label{tab:factorial_multihop}
\begin{tabular}{lcc}
\toprule
Retrieval path & Without consolidation & With consolidation \\
\midrule
Dense only & 59.57 (168) & 65.25 (184) \\
Dense + codebook & 64.89 (183) & \textbf{69.15 (195)} \\
\bottomrule
\end{tabular}
\end{table}

Table~\ref{tab:factorial_multihop} separates the codebook path from consolidation under a fixed representation and retrieval budget. The codebook path adds 5.32 points without consolidation and 3.90 points with it; consolidation adds 5.67 points for dense-only retrieval and 4.26 points for dual-path retrieval. The $-1.42$-point interaction indicates partial overlap, while each component retains a positive marginal contribution.

These ablations rule out a simple retrieval-budget explanation. More vector candidates mostly add near-duplicates from the same local embedding neighborhood, while the codebook path changes the access relation by following repeated co-occurrence among discrete anchors. The stride result gives the same message from the representation side: textual memory is dense in names, dates, quantities, and speaker states, so coarse compression removes exactly the tokens that later make temporal and compositional questions answerable. MemCode therefore gains by aligning the write unit, discrete index, and evidence reconstruction path rather than by tuning one isolated retriever. This indicates that future improvements in long-term memory should focus on enhancing the semantic structure of the stored units and the associative topology rather than merely increasing candidate counts or vector dimensions.

\begin{table}[!tbp]
\centering
\footnotesize
\caption{\label{tab:bm25_replacement_main}Fixed-budget retrieval controls on LoCoMo (GPT-4o-mini). Panel (a) uses the same frozen 5,911-AMU store and at most 40 evidence items. Panel (b) isolates the second path under a 20+20 budget; $\Delta$ is codebook minus BM25, C/B-only counts paired wins, and BM25 inj. reports additional base facts/time buckets beyond the vector stage.}
\textbf{(a) Overall same-store controls}\par\vspace{1mm}
\begin{tabular}{lcc}
\toprule
Method & Evidence budget & Overall accuracy (\%) \\
\midrule
Dense-20 & 20 & 73.18 \\
Dense-40 & 40 & 74.61 \\
Dense-20 + BM25-20 & 20+20 & 76.75 \\
Dense--sparse hybrid + reranker & 40 & 76.36 \\
HyDE + reranker~\citep{gao2023hyde} & 40 & 76.69 \\
Dense-20 + codebook-20 (MemCode) & 20+20 & \textbf{78.12} \\
\bottomrule
\end{tabular}

\vspace{1.5mm}
\textbf{(b) Codebook versus BM25 by question category}\par\vspace{1mm}
\setlength{\tabcolsep}{3.5pt}
\begin{tabular}{lccccccc}
\toprule
Category & $n$ & Codebook & BM25 & $\Delta$ & C-only & B-only & BM25 inj. \\
\midrule
Overall & 1540 & \textbf{78.12} & 76.75 & +1.36 & 105 & 84 & 18.58/10.96 \\
Multi-Hop & 282 & \textbf{64.89} & 60.64 & +4.26 & 33 & 21 & 18.50/10.65 \\
Temporal & 321 & \textbf{76.64} & 70.41 & +6.23 & 35 & 15 & 18.66/11.34 \\
Open & 96 & 43.75 & \textbf{50.00} & -6.25 & 5 & 11 & 18.59/10.78 \\
Single-Hop & 841 & 87.04 & \textbf{87.63} & -0.59 & 32 & 37 & 18.58/10.93 \\
\bottomrule
\end{tabular}
\end{table}

Table~\ref{tab:bm25_replacement_main}(a) separates retrieval structure from evidence volume. Increasing dense retrieval from 20 to 40 adds 1.43 points, while BM25, dense--sparse fusion, and HyDE remain below MemCode under the shared maximum budget.

Panel (b) then changes only the second path. BM25 injects many additional facts and time buckets, but it does not reproduce the codebook gain on Temporal and Multi-Hop questions. Sparse lexical expansion helps when surface terms already point to relevant context, whereas the codebook path helps when evidence must be joined across time, entities, and repeated discrete anchors. Appendix Tables~\ref{tab:appendix_bm25_vs_codebook} and~\ref{tab:appendix_hipporag_cost} provide the trace-level category and explicit-graph cost details.

\section{Conclusion}
Continuous nearest-neighbor retrieval leaves systematic gaps in structured long-term memory queries. Our proposed MemCode keeps the RAG-style paradigm but couples atomic abstraction with semantic quantization, topology-aware dual-path retrieval, and asynchronous consolidation. This gives facts stable, discrete anchors and exposes associations beyond vector nearest neighbors. Across LongMemEval-S and LoCoMo, MemCode achieves the best overall accuracy under two backbones while preserving construction efficiency. The component and BM25 replacement ablations further clarify the mechanism: extra lexical candidates or larger vector budgets do not recover the same temporal and multi-hop gains unless retrieval follows the discrete co-occurrence structure written during memory construction. This keeps the claim narrow: the advantage comes from matching memory writing, indexing, and reconstruction, not from increasing context volume at query time. The remaining gaps in open questions reflect a clear limitation of AMU-based abstraction: MemCode preserves explicit entities, times, speakers, and factual links well, but it may miss implicit motivations, emotions, and pragmatic cues that are better preserved in the raw dialogue.

\section*{Acknowledgments}
This research is supported by the National Key Research and Development Program of China (Grant No.2025CSJGG01800/YDZX20263100004012), the National Natural Science Foundation of China (No. 62406225), the Shanghai Municipal Commission of Economy and Informatization under the AI for Science "Hundred Teams, Hundred Projects" Program (No. 2025-GZL-RGZN-BTBX-02020), and the Fundamental Research Funds for the Central Universities.


\clearpage
{%
\bibliographystyle{unsrtnat}
\bibliography{ref}
}

\clearpage
\appendix
\section{Limitations}
\label{sec:limitations}
MemCode targets retrieval-intensive long-term conversational memory, not fully deployed agents with planning, tool use, and interactive adaptation. Its LLM-based AMU writing can miss or distort implicit motivations, emotions, and pragmatic cues; these cases explain why full-context methods remain competitive on LoCoMo Open and bound our claims to structured memory retrieval. Our HippoRAG2 comparison also measures one-shot graph construction over complete LoCoMo histories; repeated graph maintenance in a growing online memory is outside the present evaluation.

\section{Implementation Settings}

MemCode uses a lightweight local memory stack. Dialogue and atomic-memory metadata are stored in SQLite, including raw utterances, extracted AMUs, codebook IDs, and code-to-AMU links. Dense retrieval uses a FAISS L2 index over 768-dimensional bge-base-en-v1.5 embeddings~\citep{bge2024}, while the discrete path uses a 1024-entry VQ codebook and maintains code frequency and code co-occurrence statistics in the metadata store. The online stage writes AMUs asynchronously, assigns both dense vectors and discrete code indices, and reconstructs evidence at query time by combining vector retrieval with topology-aware codebook expansion. The compute environments for memory runtime and offline ISQ training are summarized in Table~\ref{tab:appendix_compute}.

\begin{table}[H]
\centering
\small
\caption{Compute environments used by MemCode. Offline ISQ codebook training is performed once and is not included in the per-dialogue memory-runtime measurements in Tables~\ref{tab:main_results} and~\ref{tab:locomo_results}.}
\label{tab:appendix_compute}
\begin{tabular}{@{}p{0.24\linewidth}p{0.26\linewidth}p{0.42\linewidth}@{}}
\toprule
Stage & Use & Hardware \\
\midrule
Memory runtime & Ingestion, retrieval, consolidation, and benchmark evaluation & 12th Gen Intel Core i5-12500H @ 3.10GHz; NVIDIA GeForce RTX 3050 Ti Laptop GPU, 4 GB; 16 GB RAM \\
Offline ISQ training & VQ-VAE codebook learning before memory evaluation & 12 vCPU Intel Xeon Platinum 8352V @ 2.10GHz; one 32 GB vGPU; 90 GB RAM \\
\bottomrule
\end{tabular}
\end{table}

Beyond the compute environments, reproducibility also depends on the fixed codebook-training configuration and the dataset-specific retrieval budgets. Table~\ref{tab:appendix_vqvae_training} summarizes the ISQ training setup, and Table~\ref{tab:appendix_implementation_budget} summarizes the dataset-specific ingestion and retrieval settings.

\begin{table}[H]
\centering
\small
\caption{Offline VQ-VAE training and ISQ codebook settings. These settings are fixed for all memory-runtime experiments.}
\label{tab:appendix_vqvae_training}
\begin{tabular}{@{}p{0.18\linewidth}p{0.28\linewidth}p{0.46\linewidth}@{}}
\toprule
Block & Item & Setting \\
\midrule
Data & Training pairs & DailyDialog context--response pairs; the context contains the two previous utterances \\
Data & Sequence length & Source and target sequences are truncated or padded to 128 tokens \\
Model & Base encoder & T5-base encoder; code dimension equals the T5 hidden size \\
Model & Projection and codebook & Stride-1 Conv1d projection, kernel size 3, padding 1; 1024 Euclidean code vectors \\
Optimization & Optimizer and schedule & AdamW, weight decay 0.01; batch size 16; gradient accumulation 2; 10 epochs; cosine schedule with 10\% warmup; gradient clipping at 1.0 \\
Optimization & Learning rates & $10^{-4}$ for T5 parameters; $10^{-3}$ for VQ, projection, and decoder-side parameters \\
Objective & Quantization loss & Generation cross-entropy plus VQ loss; commitment cost 0.10; VQ loss weight 1.0; entropy usage regularizer weight 0.01 \\
Codebook & EMA and anti-collapse & EMA decay 0.99; codebook initialized from 4096 valid encoder states; codes with EMA cluster size below 1.0 are revived every 20 optimizer steps \\
Inference & Fingerprint construction & Adjacent duplicate code ids are removed before building AMU fingerprints and co-occurrence topology \\
\bottomrule
\end{tabular}
\end{table}

\begin{table}[H]
\centering
\small
\caption{Dataset-specific ingestion and evidence-reconstruction budgets.}
\label{tab:appendix_implementation_budget}
\resizebox{\linewidth}{!}{%
\begin{tabular}{lccccc}
\toprule
Dataset & Ingestion unit & Extraction background & Dense path & Codebook path & Final evidence budget \\
\midrule
LoCoMo & A--B dialogue turn & 10 prior sentences & 20 AMUs & 20 AMUs & 40 AMUs + 25 raw dialogue turns \\
LongMemEval-S & User--Assistant turn & 5 prior sentences & 20 AMUs & 20 AMUs & 40 AMUs + 10 raw dialogue turns \\
\bottomrule
\end{tabular}
}
\end{table}

The extraction background controls how much local context is available when the system rewrites dialogue into self-contained AMUs, while the final evidence budget controls how much retrieved material is shown to the answer generator. LoCoMo uses a larger raw-context budget because image captions, speaker roles, and local conversational references often disambiguate short facts. LongMemEval-S has longer sessions and more topic drift, so its smaller raw-context budget keeps the final prompt closer to the retrieved AMUs and reduces unrelated historical evidence.

\section{Hyperparameter Sensitivity and Run Variability}

\subsection{Codebook Retrieval Hyperparameters}

We also test the sensitivity of the codebook path to three retrieval hyperparameters on LoCoMo: the number of expanded neighbors $K$, the neighbor-weight decay coefficient $\alpha$, and the IDF threshold $\tau_{\mathrm{IDF}}$. The default configuration in the main paper is $K{=}3$, $\alpha{=}0.3$, and $\tau_{\mathrm{IDF}}{=}0.5$, with an overall accuracy of 78.12\%.

\begin{table}[H]
\centering
\small
\caption{Additional hyperparameter ablations for the codebook retrieval path on LoCoMo (GPT-4o-mini). $\Delta$ denotes the signed percentage-point difference relative to the deployed default configuration $(K{=}3,\alpha{=}0.3,\tau_{\mathrm{IDF}}{=}0.5)$ with 78.12\% overall accuracy.}
\label{tab:appendix_hparam_ablation}
\begin{tabular}{llcc}
\toprule
Hyperparameter & Value & Accuracy (\%) & $\Delta$ vs.\ default \\
\midrule
\multirow{5}{*}{$K$}
& 1 & 77.40 & -0.72 \\
& \textbf{3 (default)} & \textbf{78.12} & \textbf{+0.00} \\
& 5 & 77.47 & -0.65 \\
& 7 & 77.47 & -0.65 \\
& 9 & 76.49 & -1.63 \\
\midrule
\multirow{4}{*}{$\alpha$}
& 0.1 & 77.79 & -0.33 \\
& \textbf{0.3 (default)} & \textbf{78.12} & \textbf{+0.00} \\
& 0.5 & 77.34 & -0.78 \\
& 0.7 & 77.53 & -0.59 \\
\midrule
\multirow{5}{*}{$\tau_{\mathrm{IDF}}$}
& 0.1 & 77.27 & -0.85 \\
& 0.3 & 77.34 & -0.78 \\
& \textbf{0.5 (default)} & \textbf{78.12} & \textbf{+0.00} \\
& 0.7 & 77.53 & -0.59 \\
& 0.9 & 77.40 & -0.72 \\
\bottomrule
\end{tabular}
\end{table}

Table~\ref{tab:appendix_hparam_ablation} shows modest variation across the tested range, with the default configuration consistently ahead of the alternatives. The result suggests a balance among neighbor expansion, propagation strength, and noise suppression. $K$ is the most sensitive parameter: increasing the number of expanded neighbors to 9 drops accuracy to 76.49\%, indicating that excessive expansion adds noise. Changes in $\alpha$ and $\tau_{\mathrm{IDF}}$ are smoother. The codebook path is robust to these two parameters, although moderate propagation strength and filtering give the most stable retrieval.

\subsection{Repeated-run Variability}

Although all LLM calls use temperature-0 decoding, the full hosted-LLM pipeline is not perfectly deterministic in practice. Small residual differences can arise from service-side execution, long-prompt boundary cases in memory writing and answer generation, and automatic judge decisions near the correctness boundary. We repeated the main results of LoCoMo using the GPT-4o-mini model six times, using the same retrieval configuration as Tables~\ref{tab:main_results} and~\ref{tab:locomo_results}. The run IDs in Table~\ref{tab:appendix_repeated_run_variability} are renumbered chronologically from 1 to 6; they are not tied to internal experiment-folder suffixes.

\begin{table}[H]
\centering
\footnotesize
\caption{Repeated runs for the main results of LoCoMo on GPT-4o-mini setting under temperature-0 decoding. Category columns report accuracy. The main-table result corresponds to Run 6, which is close to the center of the repeated-run distribution and is not the best run.}
\label{tab:appendix_repeated_run_variability}
\resizebox{\linewidth}{!}{%
\begin{tabular}{lcccccc}
\toprule
Run & Correct / Total & Overall & Multi-Hop & Temporal & Open & Single-Hop \\
\midrule
1 & 1200 / 1540 & 77.92 & 64.54 & 77.88 & 48.96 & 85.73 \\
2 & 1199 / 1540 & 77.86 & 64.89 & 76.95 & 50.00 & 85.73 \\
3 & 1210 / 1540 & 78.57 & 62.77 & 78.19 & 48.96 & 87.40 \\
4 & 1207 / 1540 & 78.38 & 65.25 & 76.32 & 47.92 & 87.04 \\
5 & 1212 / 1540 & 78.70 & 65.60 & 77.26 & 47.92 & 87.16 \\
6 (main table) & 1203 / 1540 & 78.12 & 64.89 & 76.64 & 43.75 & 87.04 \\
\midrule
Mean $\pm$ Std. & -- & 78.26 $\pm$ 0.35 & 64.66 $\pm$ 0.99 & 77.21 $\pm$ 0.72 & 47.92 $\pm$ 2.19 & 86.68 $\pm$ 0.75 \\
Range & -- & 77.86--78.70 & 62.77--65.60 & 76.32--78.19 & 43.75--50.00 & 85.73--87.40 \\
\bottomrule
\end{tabular}
}
\end{table}

Table~\ref{tab:appendix_repeated_run_variability} shows that the main results of GPT-4o-mini on LoCoMo are stable across repeated runs: the full overall range is 0.84 percentage points, corresponding to 13 questions out of 1540. The reported 78.12\% is slightly below the six-run mean of 78.26\% and is therefore a representative run rather than a best-case selection. Category-level variation is largest on Open because it contains only 96 questions; by contrast, the larger Multi-Hop, Temporal, and Single-Hop groups remain within narrow ranges.

\section{Backbone Sensitivity}

The main results show a consistent but counter-intuitive pattern: GPT-4o-mini is stronger than Qwen3-30B-A3B-Instruct-2507 in several MemCode settings, even though Qwen3 is a newer and generally stronger backbone. This gap is not simply a raw model-capability result. In MemCode, the backbone participates in both memory writing and answer generation, so a more verbose or more expansive model can change the density and noise profile of the memory store itself. To isolate this effect, the automatic judge is held fixed across the two backbones.

\begin{table}[H]
\centering
\footnotesize
\caption{Backbone-level performance and generation behavior for MemCode without consolidation. Category columns report accuracy. Avg.\ Ans.\ reports average answer length in words.}
\label{tab:appendix_backbone_accuracy}
\resizebox{\linewidth}{!}{%
\begin{tabular}{llcccccccc}
\toprule
Benchmark & Backbone & Overall & Temporal & Multi & Open & Single & Know.-Upd. & Single-Asst. & Avg. Ans. \\
\midrule
LoCoMo & GPT-4o-mini & \textbf{78.12} & \textbf{76.64} & \textbf{64.89} & 43.75 & 87.04 & -- & -- & 73.12 \\
LoCoMo & Qwen3 & 76.36 & 67.91 & 59.22 & \textbf{52.08} & \textbf{88.11} & -- & -- & 128.88 \\
\midrule
LongMemEval-S & GPT-4o-mini & \textbf{77.20} & \textbf{69.92} & \textbf{65.41} & -- & 92.86 & \textbf{82.05} & \textbf{92.86} & 75.52 \\
LongMemEval-S & Qwen3 & 73.80 & 57.89 & 64.66 & -- & \textbf{94.29} & 78.21 & \textbf{92.86} & 146.37 \\
\bottomrule
\end{tabular}
}
\end{table}

Table~\ref{tab:appendix_backbone_accuracy} shows that the gap is concentrated in precision-sensitive categories. On LoCoMo, GPT-4o-mini is 8.73 points higher on Temporal and 5.67 points higher on Multi, while Qwen3 is higher on Open and slightly higher on Single. On LongMemEval-S, the largest gap is again Temporal: 69.92\% for GPT-4o-mini versus 57.89\% for Qwen3. This pattern suggests that Qwen3 is not uniformly weaker. Rather, it is less stable under a fixed evidence budget when the question requires exact dates, counts, or entity binding.

\begin{table}[H]
\centering
\footnotesize
\caption{Backbone-induced memory-store characteristics. Values are averaged over dialogue-level memory stores. Qwen3 writes more AMUs and activates more codebook assignments, which increases coverage but also raises the density of potential distractors.}
\label{tab:appendix_backbone_memory_density}
\resizebox{\linewidth}{!}{%
\begin{tabular}{llccccc}
\toprule
Benchmark & Backbone & AMUs / dialog & Active codes & Code assignments & Max code freq. & Avg. AMU words \\
\midrule
LoCoMo & GPT-4o-mini & 591.1 & 216.9 & 3678.6 & 294.0 & 16.48 \\
LoCoMo & Qwen3 & 940.4 & 266.1 & 6150.2 & 461.4 & 15.92 \\
\midrule
LongMemEval-S & GPT-4o-mini & 1553.4 & 420.4 & 9941.1 & 798.0 & 21.41 \\
LongMemEval-S & Qwen3 & 2130.3 & 450.2 & 14532.2 & 1356.1 & 21.80 \\
\bottomrule
\end{tabular}
}
\end{table}

The memory statistics in Table~\ref{tab:appendix_backbone_memory_density} explain why the stronger backbone does not automatically yield higher end-to-end memory accuracy. Qwen3 produces substantially more AMUs: 59.1\% more on LoCoMo and 37.1\% more on LongMemEval-S. It also produces many more code assignments, increasing by 67.2\% and 46.2\% respectively. This denser write behavior can improve coverage for open-ended questions, but it also creates more near-duplicate facts and more competing codebook activations. Since the answer prompt still keeps a fixed number of AMUs, higher memory density can make the selected evidence less clean.

\begin{table}[H]
\centering
\small
\caption{Paired question-level disagreements between GPT-4o-mini and Qwen3 under MemCode. \textit{GPT-only} means GPT-4o-mini answers correctly while Qwen3 does not; \textit{Qwen-only} is the reverse.}
\label{tab:appendix_backbone_pairwise}
\resizebox{\linewidth}{!}{%
\begin{tabular}{lcccccc}
\toprule
Benchmark & Shared $n$ & Both correct & Both wrong & GPT-only & Qwen-only & Largest GPT-only source \\
\midrule
LoCoMo & 1540 & 1055 & 216 & 148 & 121 & Temporal: 55 vs. 27 \\
LongMemEval-S & 500 & 322 & 67 & 64 & 47 & Temporal: 28 vs. 12 \\
\bottomrule
\end{tabular}
}
\end{table}

The paired comparison in Table~\ref{tab:appendix_backbone_pairwise} supports the same interpretation. GPT-4o-mini has more unique wins on both benchmarks, and the largest asymmetry comes from temporal reasoning. Manual inspection of the disagreement patterns shows recurring Qwen3 failure modes: answering with the total amount rather than the per-item amount, selecting a nearby but wrong event date, over-explaining abstention questions, or following a plausible distractor when multiple related facts are retrieved. These errors are consistent with a memory-system interaction: a more expansive writer and generator improves recall coverage, but it also increases distractor density and makes exact evidence selection harder. The backbone gap therefore reflects how the model interacts with the external memory pipeline, not a simple ranking of model capability in isolation.

\section{Further Retrieval and Codebook Analysis}

This appendix examines retrieval-path behavior through same-budget controls, dense-embedding sensitivity, question-level novelty analysis, case studies, global codebook usage, and a local visualization of semantic purity.

\subsection{Codebook Path vs.\ a BM25 Keyword Second Path}

The main paper summarizes a second-path replacement ablation in Table~\ref{tab:bm25_replacement_main}. Here we provide the full trace-level version. We replace MemCode's codebook path with vanilla BM25 keyword retrieval under the same ``20+20'' retrieval budget: Path A remains dense vector retrieval, while Path B changes from discrete codebook retrieval to BM25.

\begin{table}[H]
\centering
\footnotesize
\caption{Full details for the second-path replacement ablation on LoCoMo (GPT-4o-mini). $\Delta$ denotes codebook minus BM25. ``C-only'' counts cases where the codebook path answers correctly while BM25 does not, and ``B-only'' denotes the reverse. ``New Base'' and ``New Time'' report the average number of additional base facts and time buckets injected by the BM25 stage beyond the vector stage.}
\label{tab:appendix_bm25_vs_codebook}
\resizebox{\linewidth}{!}{%
\begin{tabular}{lcccccccc}
\toprule
\multirow{2}{*}{Category} & \multirow{2}{*}{$n$} & \multicolumn{3}{c}{Accuracy} & \multicolumn{2}{c}{Paired Wins} & \multicolumn{2}{c}{BM25 Injection} \\
\cmidrule(lr){3-5}\cmidrule(lr){6-7}\cmidrule(lr){8-9}
 &  & Codebook & BM25 & $\Delta$ & C-only & B-only & New Base & New Time \\
\midrule
Overall & 1540 & \textbf{78.12} & 76.75 & +1.36 & 105 & 84 & 18.58 & 10.96 \\
Multi-Hop & 282 & \textbf{64.89} & 60.64 & +4.26 & 33 & 21 & 18.50 & 10.65 \\
Temporal Reasoning & 321 & \textbf{76.64} & 70.41 & +6.23 & 35 & 15 & 18.66 & 11.34 \\
Open Domain & 96 & 43.75 & \textbf{50.00} & -6.25 & 5 & 11 & 18.59 & 10.78 \\
Single-Hop / Basic & 841 & 87.04 & \textbf{87.63} & -0.59 & 32 & 37 & 18.58 & 10.93 \\
\bottomrule
\end{tabular}
\vspace{1mm}
}
\end{table}

The codebook path is not directly replaceable by a lexical second path. Overall accuracy rises from 76.75\% with BM25 to 78.12\% with the codebook path, or 21 more correct answers. The advantage is concentrated in the categories closest to structured memory retrieval: Temporal Reasoning improves by 6.23 percentage points and Multi-Hop by 4.26 points. BM25 is 6.25 points higher on Open Domain, but that category contains only 96 questions; on Single-Hop / Basic, BM25 has only a 0.59-point edge. The distribution matches the inductive biases of the two paths. BM25 favors surface term overlap, whereas the codebook path helps when evidence has stronger temporal structure or higher compositionality.

The paired comparison over 1540 shared question-level traces supports the category-level result. There are 105 samples where the codebook path is correct and BM25 is wrong, compared with 84 in the reverse direction. The asymmetry again comes from the more complex categories. On Temporal Reasoning, the codebook path has 35 ``codebook-only wins'' versus 15 ``BM25-only wins''; on Multi-Hop, the counts are 33 versus 21. BM25 shows only local advantages on Single-Hop / Basic (37 versus 32) and Open Domain (11 versus 5).

Retrieval traces also show that BM25 is not empty expansion: relative to the vector stage, it injects about 18.6 additional new base facts and about 11 new time buckets per question on average. Still, this lexical expansion does not close the gap on temporal and compositional questions. The codebook path helps by adding candidates that better match structured memory retrieval, rather than by merely expanding the candidate set.

These results should not be read as showing that the codebook path is uniformly better than BM25 across all question types. Under the same dual-path budget, BM25 is a reasonable sparse-retrieval control, but it does not reproduce the gain pattern of the codebook path, especially its relative advantage on temporal and multi-hop questions.

\subsection{Same-Budget Retrieval Controls and an Explicit-Graph Baseline}

As a supplementary explicit-graph control, we evaluate Full HippoRAG2~\citep{gutierrez2025hipporag2} on the same frozen 5,911-AMU store with at most 40 evidence items per query. It answers 1,200 of 1,540 questions correctly, compared with 1,203 for MemCode; the 0.20-point difference is not statistically significant in the paired analysis. Thus, an explicit graph can recover comparable associations, while Table~\ref{tab:appendix_hipporag_cost} isolates the additional construction and query profile.

\begin{table}[H]
\centering
\footnotesize
\caption{Additional retrieval-system cost beyond the shared AMU construction stage. Query latency is averaged over LoCoMo questions.}
\label{tab:appendix_hipporag_cost}
\resizebox{0.90\linewidth}{!}{%
\begin{tabular}{lcc}
\toprule
Metric & MemCode & Full HippoRAG2@40 \\
\midrule
Additional model calls & 0 & 12,779 \\
Additional tokens & 0 & 8.47M \\
Explicit graph construction & None & 34.49 min \\
Graph size & None & 17,275 nodes / 109,779 edges \\
Query latency & 0.060 s & 2.864 s \\
\bottomrule
\end{tabular}%
}
\end{table}

The explicit graph baseline is evaluated favorably with a one-shot build over each complete dialogue history. Even in this static setting, its additional construction calls and 47.8$\times$ query latency contrast with MemCode's use of code assignments already produced during AMU indexing. This comparison supports an efficiency claim, rather than a claim that graph retrieval is intrinsically less accurate.

\subsection{Sensitivity to the Dense Embedding Model}
\label{sec:appendix_embedding_sensitivity}

We further test whether MemCode's LoCoMo result depends on the particular dense embedding backbone used by the vector path. The main experiments use bge-base-en-v1.5. As a smaller and widely used alternative, we replace the dense encoder with all-MiniLM-L6-v2 while keeping the GPT-4o-mini writer/reader, the 1024-entry codebook, and the same ``20+20'' retrieval budget fixed. Table~\ref{tab:appendix_embedding_sensitivity} reports paired question-level results over the same 1540 LoCoMo questions.

\begin{table}[H]
\centering
\footnotesize
\caption{Embedding-model sensitivity on LoCoMo (GPT-4o-mini). The bge-base-en-v1.5 row corresponds to the main MemCode setting. $\Delta$ is the signed percentage-point difference of all-MiniLM-L6-v2 minus bge-base-en-v1.5. ``bge-only'' and ``MiniLM-only'' count paired question-level wins.}
\label{tab:appendix_embedding_sensitivity}
\resizebox{\linewidth}{!}{%
\begin{tabular}{lccccccc}
\toprule
Category & $n$ & bge Acc. & MiniLM Acc. & $\Delta$ & bge-only & MiniLM-only & Shared correct \\
\midrule
Overall & 1540 & \textbf{78.12} & 76.23 & -1.88 & 90 & 61 & 1113 \\
Multi-Hop & 282 & \textbf{64.89} & 62.77 & -2.13 & 23 & 17 & 160 \\
Temporal Reasoning & 321 & \textbf{76.64} & 73.83 & -2.80 & 20 & 11 & 226 \\
Open Domain & 96 & 43.75 & 43.75 & +0.00 & 4 & 4 & 38 \\
Single-Hop / Basic & 841 & \textbf{87.04} & 85.37 & -1.66 & 43 & 29 & 689 \\
\bottomrule
\end{tabular}
}
\end{table}

The smaller encoder preserves the qualitative result but slightly weakens retrieval accuracy. Overall accuracy decreases from 78.12\% to 76.23\%, and the paired comparison gives 90 bge-only wins versus 61 MiniLM-only wins. The drop is concentrated in the categories that require tighter evidence binding: Temporal Reasoning falls by 2.80 points and Multi-Hop by 2.13 points. Open Domain is unchanged, which is consistent with the earlier BM25 control: open-ended questions are less dominated by exact nearest-neighbor geometry and more affected by whether the final prompt contains broadly relevant context.

The memory-store statistics are nearly unchanged by embedding replacement. Averaged over the 10 LoCoMo dialogue stores, bge-base-en-v1.5 writes 591.1 AMUs, activates 216.9 codes, and produces 3678.6 code assignments per dialogue; all-MiniLM-L6-v2 writes 588.3 AMUs, activates 215.5 codes, and produces 3651.9 assignments. The performance gap therefore does not come from a different memory volume. It more likely reflects the precision of dense evidence selection before codebook expansion. This supports a conservative interpretation: MemCode does not rely on bge-base-en-v1.5 to function, but stronger dense embeddings improve the exact-evidence categories most sensitive to retrieval precision.

\subsection{Dual-Path Retrieval: Novelty Analysis and Case Studies}

\begin{figure}[H]
\centering
\includegraphics[width=\linewidth]{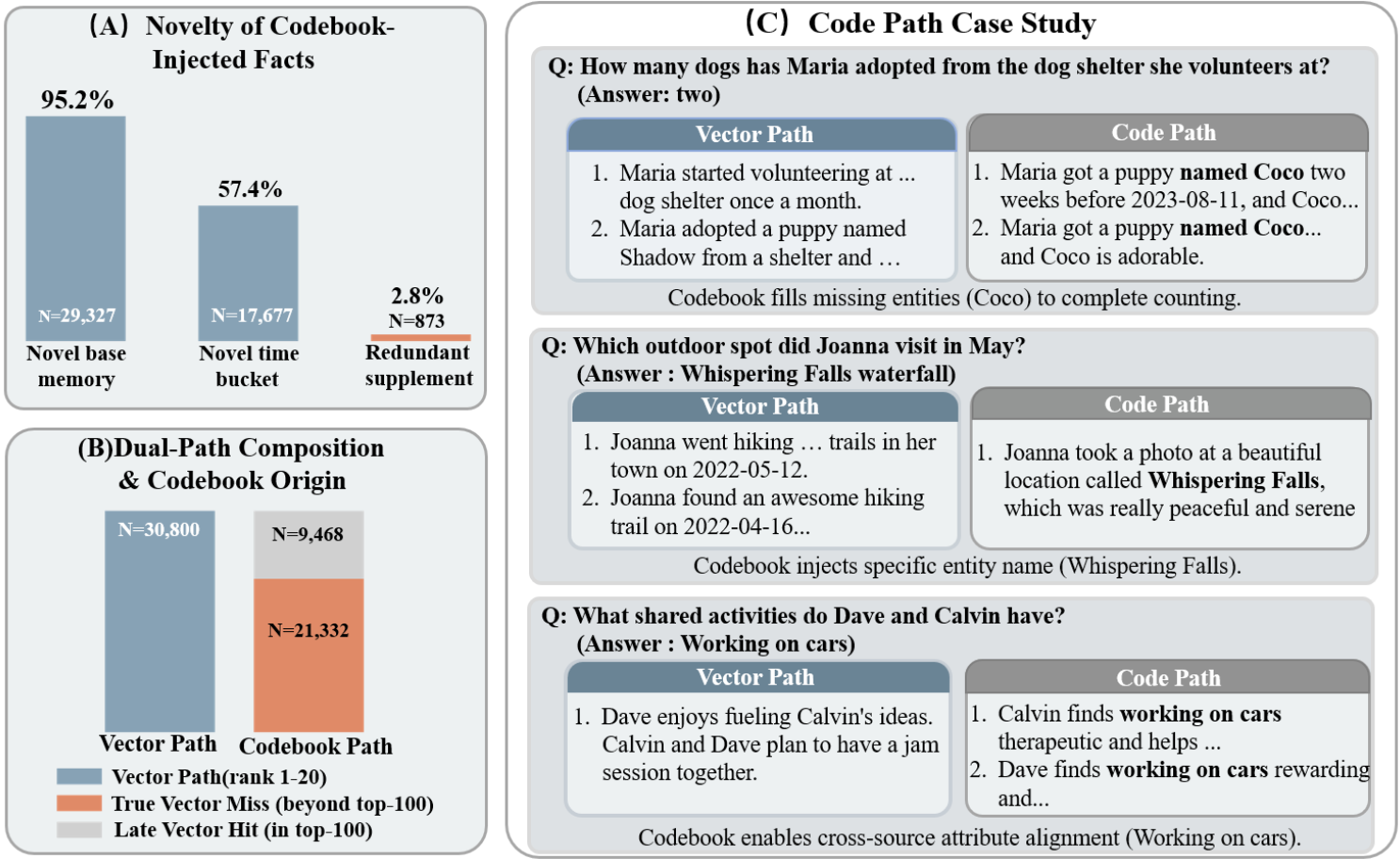}
\vspace{-2mm}
\caption{\label{fig:case_study}Dual-path retrieval: novelty analysis and case studies.}
\vspace{-5mm}
\end{figure}

We measure what the codebook second path adds by computing the proportion of facts independently hit by the codebook path but not covered by the vector path. Figure~\ref{fig:case_study} shows that the added content is closer to facts that the vector path structurally misses than to candidates the vector path could already find at lower ranks. In particular, 69.3\% of facts added by the codebook path are still absent from vector top-100, while simply doubling vector Top-$K$ brings only a 1.4\% marginal gain. The difference between the two paths is therefore not retrieval depth but representation: the vector path favors fragments with similar surface semantics, while the codebook path uses co-occurrence topology to activate memories that are logically related but more distant on the surface.

The three cases in Figure~\ref{fig:case_study}(C) show the same complementarity in concrete tasks. The failures follow different patterns: in the counting question, the vector path misses another independent entity (Coco); in the location question, it misses a low-frequency proper noun (Whispering Falls); in the shared-attribute question, it misses parallel facts scattered across dialogue fragments. Entity omission, suppression of low-frequency proper nouns, and cross-fragment attribute dispersion all follow from geometric nearest-neighbor bias. Codebook topology compensates for them by modeling co-occurrence relations. Together with the BM25 control, this novelty evidence supports a more specific claim: the codebook path recovers structured evidence that both vector and lexical paths often miss.

\subsection{Additional Codebook Space Analysis}

We also analyze the codebook space learned in the main experiments. The analysis uses metadata statistics from 10 dialogue buckets in the LoCoMo main run and focuses on sparsity, cross-session stability, global concentration, and co-occurrence topology. The 1024-dimensional codebook activates 216.9 codes on average within a single dialogue (21.18\%). The union of 10 dialogues covers 525 codes (51.27\%), while only 61 codes (5.96\%) appear in every dialogue. The learned discrete space therefore does not collapse into a flat structure where all codes are used uniformly. It forms a backbone-plus-tail pattern with a stable core and conversation-specific tail.

\begin{figure}[H]
\centering
\includegraphics[width=0.74\linewidth]{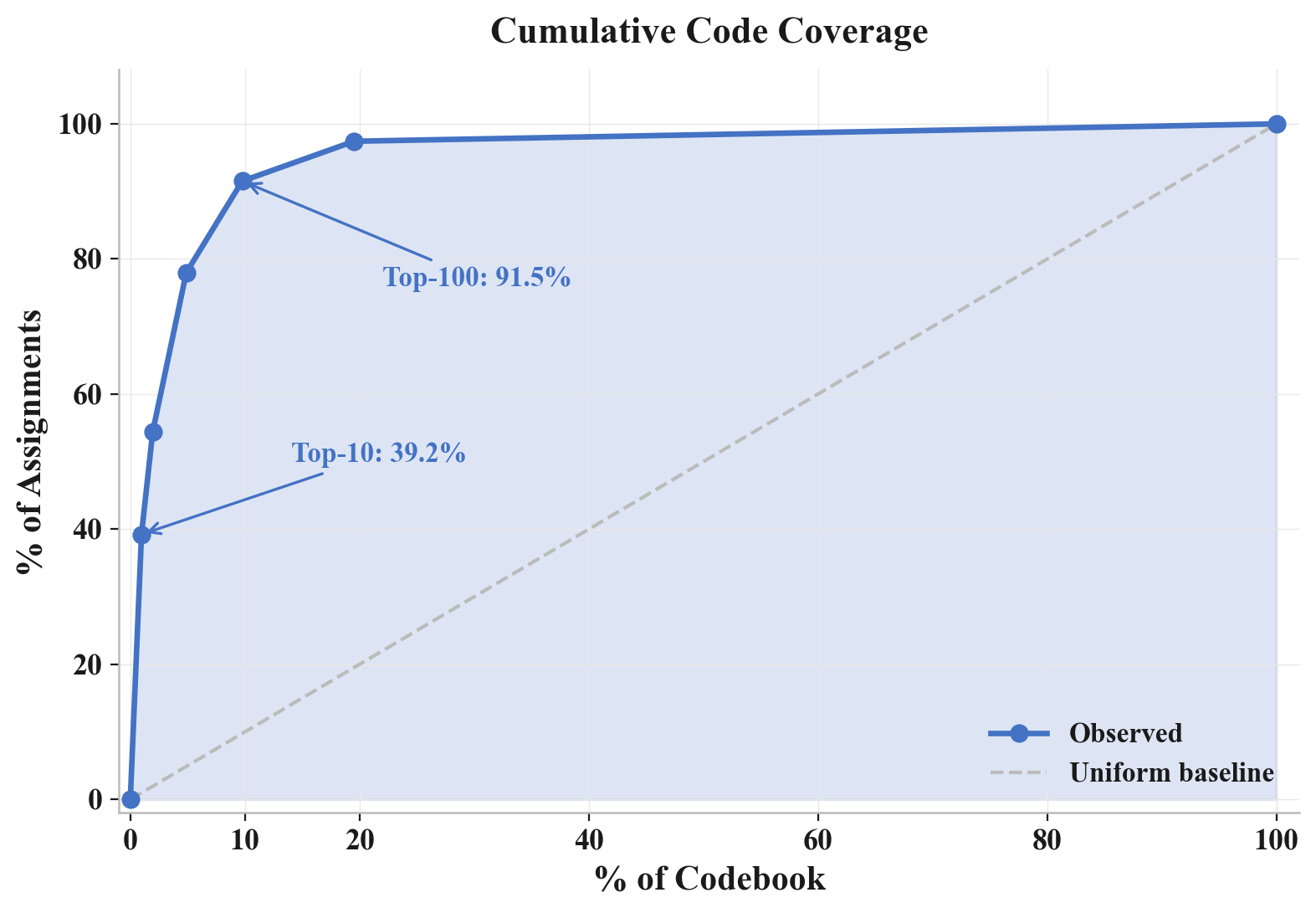}
\caption{\label{fig:appendix_codebook_usage}Global codebook usage concentration. The cumulative coverage curve over ranked codes shows the top-10 codes already explain 39.23\% of all assignments and the top-100 explain 91.53\%.}
\end{figure}

Figure~\ref{fig:appendix_codebook_usage} shows concentrated rather than uniform code usage: the top-10 codes explain 39.23\% of all AMU assignments, the top-20 explain 54.41\%, and the top-100 explain 91.53\%. This concentration is not trivial collapse, as the activated vocabulary extends well beyond a tiny fixed set of symbols.

\begin{figure}[H]
\centering
\includegraphics[width=0.52\linewidth]{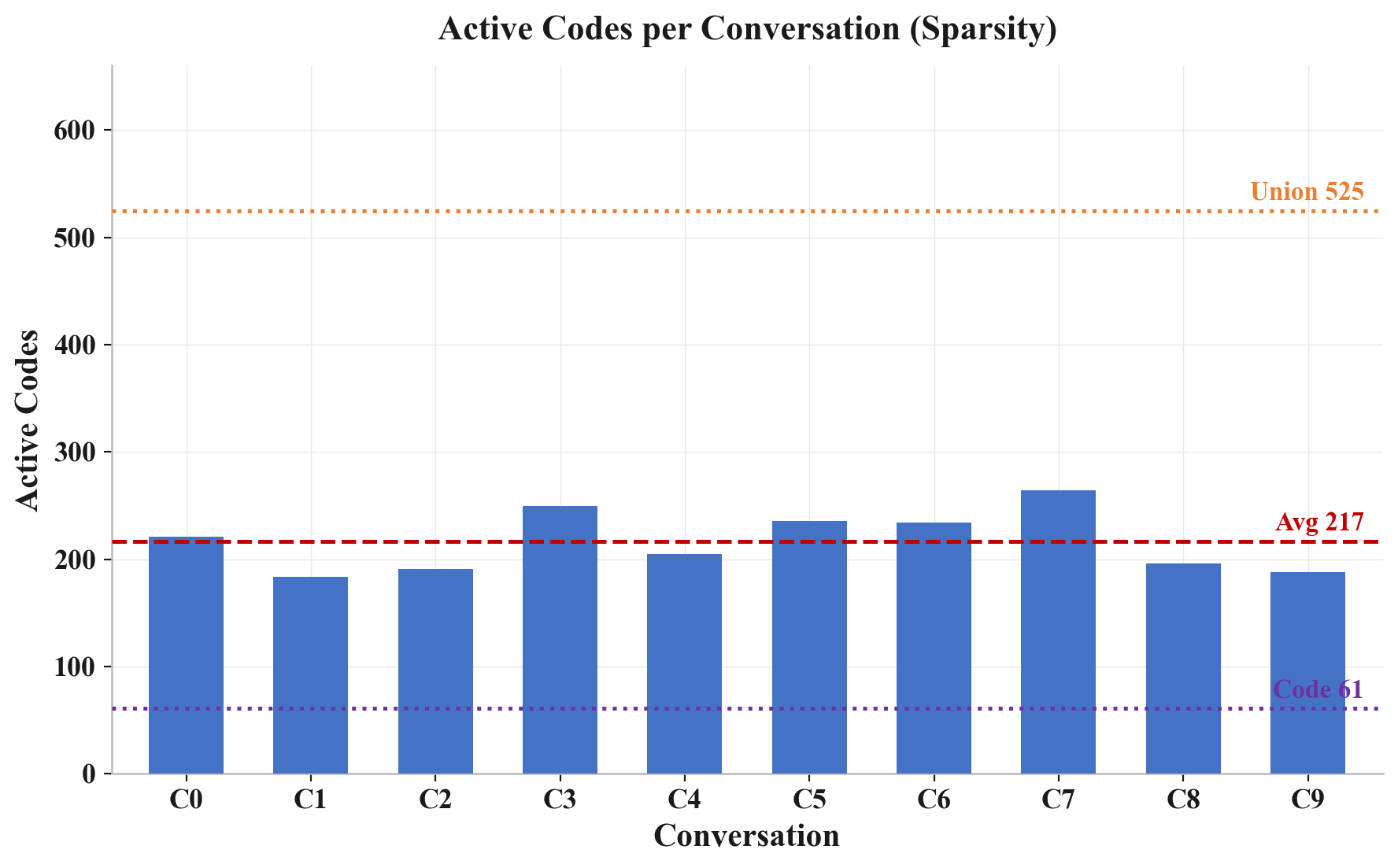}
\hfill
\includegraphics[width=0.43\linewidth]{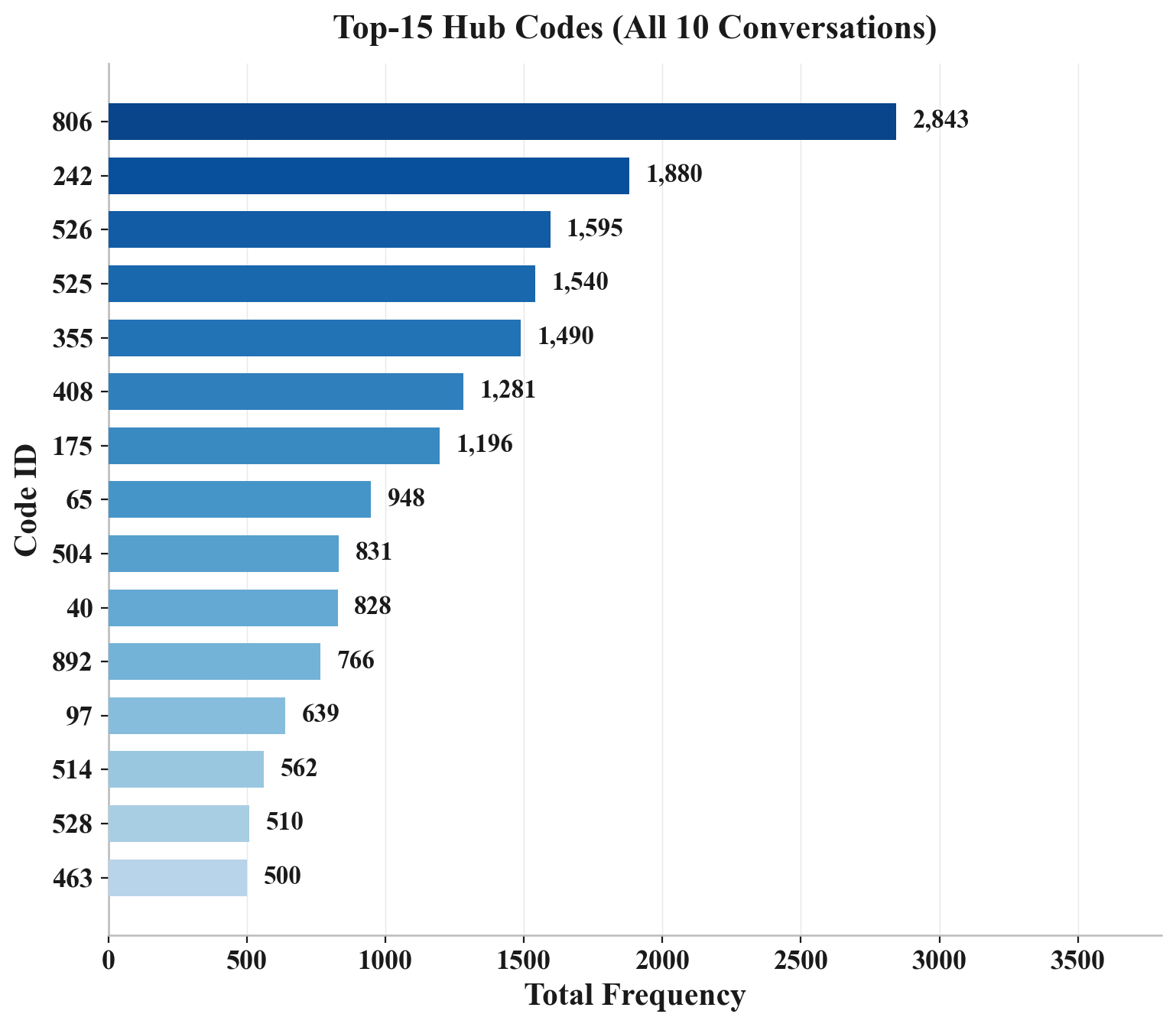}
\caption{\label{fig:appendix_codebook_sparsity_hubs}Conversation-level sparsity and shared hub codes. Left: active-code counts per conversation, showing strong per-dialogue sparsity and a small shared core. Right: frequency distribution of the top-15 hub codes across all 10 conversations; all 15 appear in every conversation, with Code 806 being the dominant hub.}
\end{figure}

Figure~\ref{fig:appendix_codebook_sparsity_hubs} complements the aggregate coverage view by showing both sides of the activation pattern. Each dialogue occupies only a small local subset of the 1024-code vocabulary, while the top hub codes recur across all 10 conversations, with Code 806 standing out as the most frequent anchor. Together, Figures~\ref{fig:appendix_codebook_usage} and~\ref{fig:appendix_codebook_sparsity_hubs} support a \emph{shared backbone + conversation-specific tail} regime, which is the setting in which a global co-occurrence topology can support associative retrieval beyond nearest-neighbor matching.

\begin{figure}[H]
\centering
\includegraphics[width=0.48\linewidth]{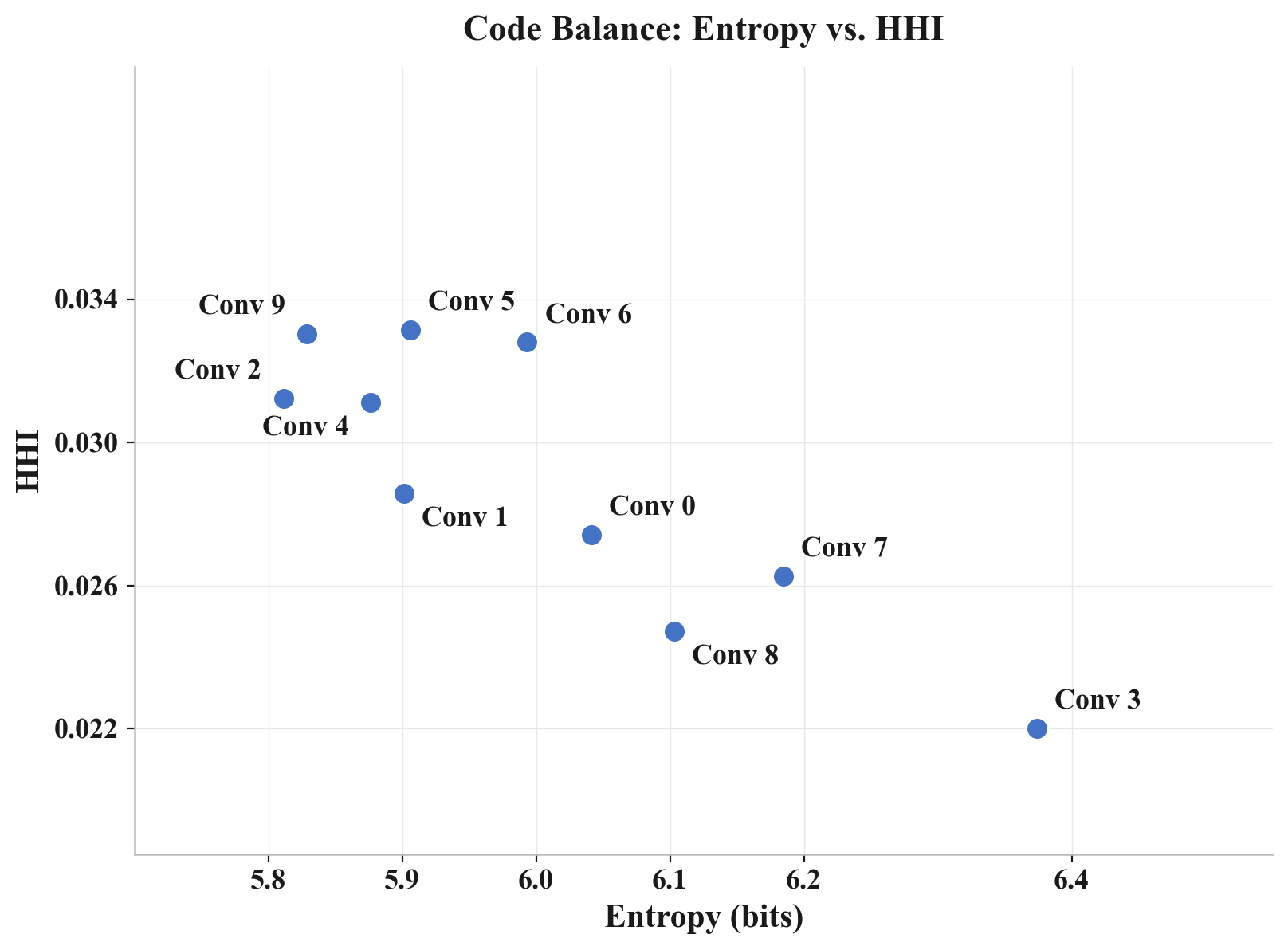}
\hfill
\includegraphics[width=0.48\linewidth]{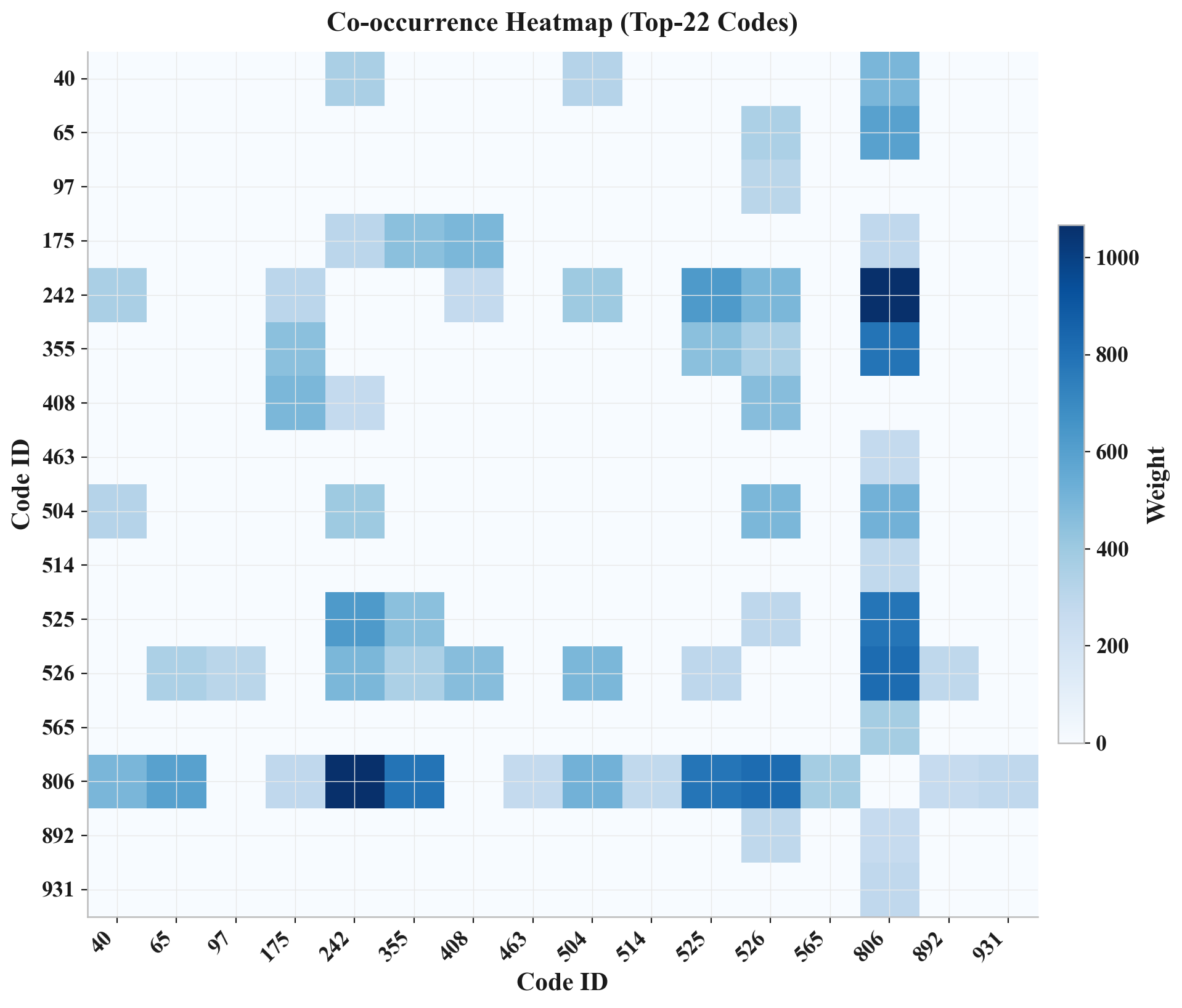}
\caption{\label{fig:appendix_codebook_topology}Codebook stability and co-occurrence topology. Left: entropy--HHI scatter across conversations; all points lie in a high-entropy, low-HHI regime, indicating that the codebook does not collapse to a few trivial symbols. Right: co-occurrence heatmap of representative high-frequency codes; the strongest pair is $(242, 806)$ with weight 1067, and visible sub-clusters such as $(242, 525, 526)$ indicate structured semantic topology.}
\vspace{-3mm}
\end{figure}

Figure~\ref{fig:appendix_codebook_topology} checks two properties of the learned discrete space. First, all conversations lie in a high-entropy ($\approx 5.8$--$6.4$ bits), low-HHI ($\approx 0.022$--$0.034$) region, with average entropy 6.00 and average HHI 0.029. This pattern is inconsistent with codebook collapse and supports the entropy regularization and dead-code revival mechanism used in ISQ training. Second, the co-occurrence heatmap is structured: Code 806 acts as a clear hub, the strongest pair $(242, 806)$ reaches weight 1067, and several sub-clusters are visibly denser than the background. Together, these observations support the claim that the codebook path provides a structured semantic topology rather than a bag of arbitrary discrete tokens.

\begin{figure}[H]
\centering
\includegraphics[width=\linewidth]{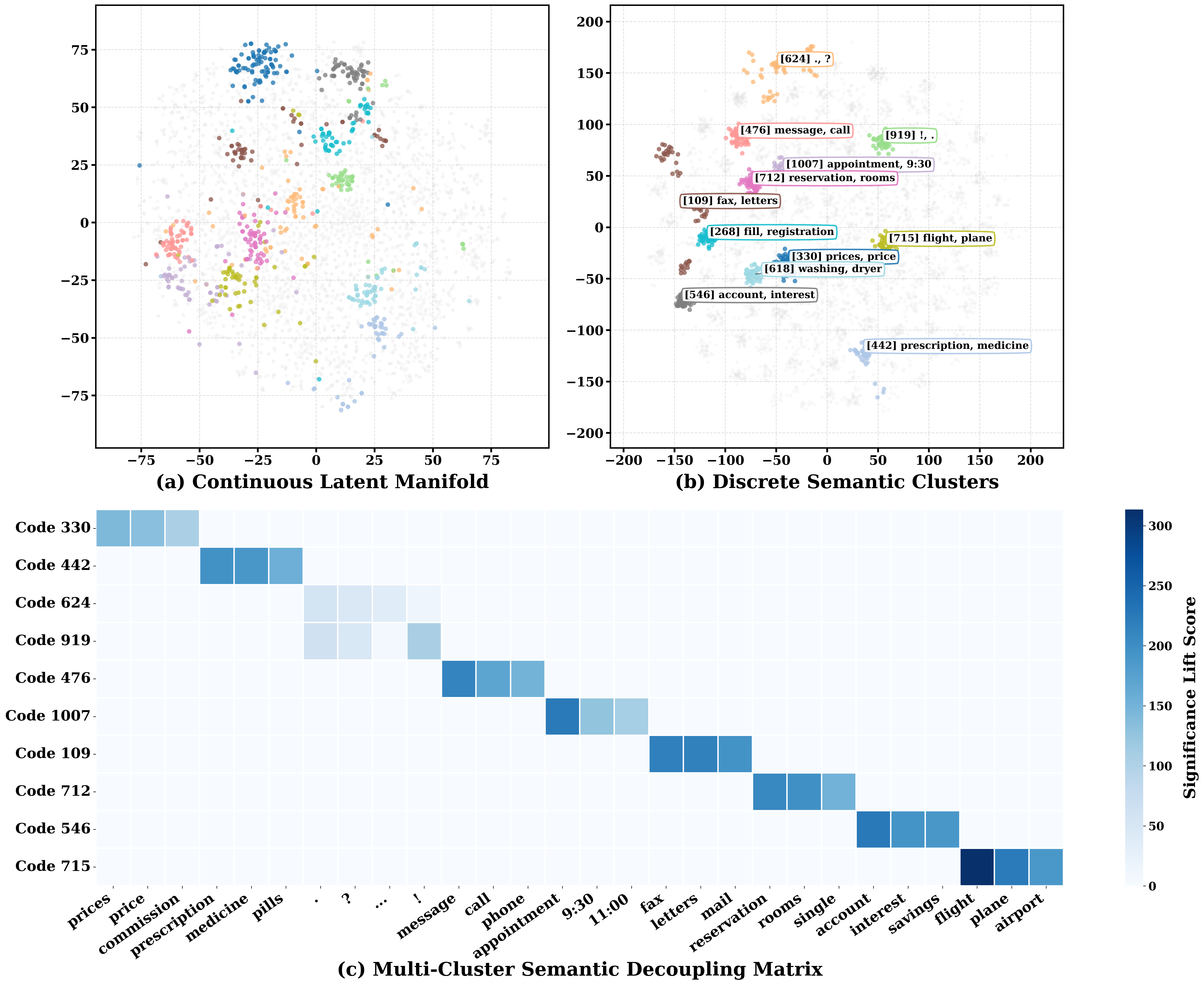}
\vspace{-3mm}
\caption{\label{fig:discrete_vis}Visualization of discrete semantic disentanglement. (a) shows the continuous latent manifold from the T5-base encoder; (b) shows topological tightening after commitment loss; (c) shows the disentanglement matrix with exclusive activation of top-$K$ codebook IDs by semantic clusters.}
\vspace{-3mm}
\end{figure}

\subsection{Visualization of Discrete Semantic Disentanglement}
To complement the aggregate codebook statistics, we inspect the geometry of the learned latent space. Figure~\ref{fig:discrete_vis}(a) shows that the continuous latent manifold produced by the T5-base encoder remains diffuse, with soft semantic boundaries before quantization. After the commitment constraint is imposed, Figure~\ref{fig:discrete_vis}(b) shows the latent space contracting into tighter and more separated discrete semantic clusters.

We probe semantic purity by analyzing the most frequently activated Top-$K$ codebook indices on the validation set. Using Statistical Significance Lift to map input tokens to these high-frequency indices, the disentanglement matrix in Figure~\ref{fig:discrete_vis}(c) has a near-diagonal sparse structure, indicating separated discrete representations~\citep{higgins2017betavae}. For example, Code 442 is selectively activated by medical terms such as ``prescription'' and ``medicine,'' while Codes 712 and 715 consistently anchor accommodation- and transportation-related contexts, respectively, with little overlap. This visualization is consistent with the broader appendix evidence: the codebook is not collapsed, and its discrete anchors are semantically meaningful.

\section{Prompt Templates}
\label{sec:appendix_prompts}

We include the core prompt templates used for memory writing and automatic evaluation. The templates are shown in normalized form with variable placeholders; dataset-specific values such as timestamps, speakers, retrieved evidence, rubrics, and reference answers are substituted at runtime.

\subsection{Atomic Memory Writing}

\begin{promptbox}{Atomic Fact Extraction with Background Context}
\small
\begingroup
\setlength{\parindent}{0pt}
\setlength{\parskip}{0.25em}
You are an expert Memoryist. Extract dense, stand-alone facts from the conversation.

\textbf{Rules}
\begin{enumerate}
\item \textbf{Filter Noise}: Ignore greetings, short reactions, and vague chitchat.
\item \textbf{Resolve Identity}: Replace pronouns (I/you/he) with actual speaker names.
\item \textbf{Time Rewriting}: Rewrite relative time terms (e.g., ``yesterday'', ``last week'') into exact dates/times. Use the timestamp header above the text as the anchor.
\item \textbf{Merge}: Combine fragmented sentences into one complete fact.
\item \textbf{Context}: If a line is marked as \texttt{[Background]}, use it for understanding but do not extract facts from it.
\end{enumerate}

\textbf{Example}
\begin{quote}
\footnotesize\ttfamily
--- Timestamp: 2023-05-10 14:30 ---\\
{[}User{]}: I just arrived.\\[0.25em]
Output Fact: ``User arrived at 14:30 on 2023-05-10.''
\end{quote}

\textbf{Input}
\begin{quote}
\footnotesize\ttfamily
\{context\}
\end{quote}

\textbf{Output strictly JSON}
\begin{quote}
\footnotesize\ttfamily
[\\
\hspace*{1em}\{"fact": "SpeakerName action...",\\
\hspace*{1.6em}"source\_indices": [1, 2]\}\\
]
\end{quote}
\endgroup
\end{promptbox}

\begin{promptbox}{Atomic Fact Extraction without Background Context}
\small
\begingroup
\setlength{\parindent}{0pt}
\setlength{\parskip}{0.25em}
You are an expert Memoryist. Extract dense, stand-alone facts from the conversation.

\textbf{Rules}
\begin{enumerate}
\item \textbf{Filter Noise}: Ignore greetings, short reactions, and vague chitchat.
\item \textbf{Resolve Identity}: Replace pronouns (I/you/he) with actual speaker names.
\item \textbf{Time Rewriting}: Rewrite relative time terms into exact dates/times using the timestamp header above the text as the anchor.
\item \textbf{Merge}: Combine fragmented sentences into one complete fact.
\end{enumerate}

\textbf{Input}
\begin{quote}
\footnotesize\ttfamily
\{context\}
\end{quote}

\textbf{Output strictly JSON}
\begin{quote}
\footnotesize\ttfamily
[\\
\hspace*{1em}\{"fact": "SpeakerName action...",\\
\hspace*{1.6em}"source\_indices": [1, 2]\}\\
]
\end{quote}
\endgroup
\end{promptbox}

The background-aware version is used when the ingestion chunk contains carry-over context from previous turns. This keeps pronoun and temporal resolution stable without allowing old context to create duplicate AMUs. The background-free version is used when the chunk itself already contains the target evidence.

\subsection{LLM-as-Judge}

\begin{promptbox}{Standard Tasks}
\small
\begingroup
\setlength{\parindent}{0pt}
\setlength{\parskip}{0.25em}
I will give you a question, a correct answer, and a response from a model. Please answer yes if the response contains the correct answer. Otherwise, answer no. If the response is equivalent to the correct answer or contains all the intermediate steps to get the correct answer, you should also answer yes. If the response only contains a subset of the information required by the answer, answer no.

\textbf{Fields}
\begin{quote}
\footnotesize\ttfamily
Reference Time: \{query\_date\} \quad(optional)\\
Question: \{question\}\\
Correct Answer: \{correct\_answer\}\\
Model Response: \{model\_response\}
\end{quote}

\textbf{Decision}
\begin{quote}
\footnotesize
Is the model response correct? Answer yes or no only.
\end{quote}
\endgroup
\end{promptbox}

\begin{promptbox}{Temporal Reasoning Tasks}
\small
\begingroup
\setlength{\parindent}{0pt}
\setlength{\parskip}{0.25em}
You are an intelligent judge evaluating the correctness of an AI's answer regarding time and dates.

\textbf{Fields}
\begin{quote}
\footnotesize\ttfamily
Reference Time: \{query\_date\} \quad(optional)\\
Question: \{question\}\\
Ground Truth: \{correct\_answer\}\\
Model Answer: \{model\_response\}
\end{quote}

\textbf{Evaluation Rules}
\begin{enumerate}
\item Date formats are equivalent when they refer to the same date.
\item If the ground truth is a simple time and the model answer contains that time in a full sentence, mark it correct.
\item If the ground truth provides multiple valid options, the model answer is correct if it matches any one of them.
\item For duration questions, allow inclusive/exclusive counting ambiguity and explicit $\pm 1$ unit differences.
\item If the answer is relative to the reference time and matches the ground-truth logic, mark it correct.
\end{enumerate}

\textbf{Decision}
\begin{quote}
\footnotesize
Does the model answer factually match the ground truth or one of its valid options? Respond with only ``yes'' or ``no''.
\end{quote}
\endgroup
\end{promptbox}

\begin{promptbox}{Knowledge Update Tasks}
\small
\begingroup
\setlength{\parindent}{0pt}
\setlength{\parskip}{0.25em}
I will give you a question, a correct answer, and a response from a model. Please answer yes if the response contains the correct answer. Otherwise, answer no. If the response contains previous information along with an updated answer, the response should be considered correct as long as the updated answer is the required answer.

\textbf{Fields}
\begin{quote}
\footnotesize\ttfamily
Reference Time: \{query\_date\} \quad(optional)\\
Question: \{question\}\\
Correct Answer: \{correct\_answer\}\\
Model Response: \{model\_response\}
\end{quote}

\textbf{Decision}
\begin{quote}
\footnotesize
Is the model response correct? Answer yes or no only.
\end{quote}
\endgroup
\end{promptbox}

\begin{promptbox}{Preference Tasks}
\small
\begingroup
\setlength{\parindent}{0pt}
\setlength{\parskip}{0.25em}
I will give you a question, a rubric for the desired personalized response, and a response from a model. Please answer yes if the response satisfies the desired response. The model does not need to reflect all points in the rubric; the response is correct as long as it recalls and uses the user's personal information correctly.

\textbf{Fields}
\begin{quote}
\footnotesize\ttfamily
Reference Time: \{query\_date\} \quad(optional)\\
Question: \{question\}\\
Rubric: \{rubric\}\\
Model Response: \{model\_response\}
\end{quote}

\textbf{Decision}
\begin{quote}
\footnotesize
Is the model response correct? Answer yes or no only.
\end{quote}
\endgroup
\end{promptbox}

\begin{promptbox}{Abstention Tasks}
\small
\begingroup
\setlength{\parindent}{0pt}
\setlength{\parskip}{0.25em}
I will give you an unanswerable question, an explanation, and a response from a model. Please answer yes if the model correctly identifies the question as unanswerable. The model could say that the information is incomplete, or that some information is given but the asked information is not.

\textbf{Fields}
\begin{quote}
\footnotesize\ttfamily
Reference Time: \{query\_date\} \quad(optional)\\
Question: \{question\}\\
Explanation: \{explanation\}\\
Model Response: \{model\_response\}
\end{quote}

\textbf{Decision}
\begin{quote}
\footnotesize
Does the model correctly identify the question as unanswerable? Answer yes or no only.
\end{quote}
\endgroup
\end{promptbox}

\section{Responsible Use and Existing Assets}
\label{sec:appendix_responsible_assets}

\textbf{Broader impacts.}
MemCode is intended to make long-term agent memory more reliable, auditable, and efficient. Better memory systems can improve personalized assistance, reduce repeated user effort across sessions, and make retrieved evidence easier to inspect through source pointers and structured memory records. The same capability also raises risks when deployed on real user histories: a persistent memory store may retain sensitive personal information, propagate extraction mistakes into future personalization, or retrieve outdated facts after user preferences or circumstances change. These risks are especially relevant because MemCode targets cross-session memory rather than isolated question answering.

Responsible deployment should therefore treat the memory store as user data rather than as ordinary model cache. Practical safeguards include explicit user consent for memory creation, user-visible inspection and deletion controls, retention limits, access control for stored memories and source dialogue, and deployment-time filters for sensitive categories. MemCode's source pointers, archived-fact records, and conservative consolidation gates provide auditability and reduce irreversible rewriting, but they do not replace product-level privacy controls or jurisdiction-specific compliance review.

\textbf{Existing assets and licenses.}
Table~\ref{tab:appendix_assets_licenses} summarizes the main third-party assets used in the experiments. We cite the original papers or release pages for these assets and use them according to their accompanying licenses or terms of use. We do not redistribute third-party datasets, model weights, or API outputs as a new dataset; any released code or supplemental material should instruct users to obtain third-party assets from their original sources and follow the corresponding license, non-commercial restriction, or service terms.

\begin{table}[H]
\centering
\small
\caption{Existing datasets, models, APIs, and software assets used by MemCode. License and terms information follows the public release pages associated with the cited assets.}
\label{tab:appendix_assets_licenses}
\resizebox{\linewidth}{!}{%
\begin{tabular}{p{0.20\linewidth}p{0.28\linewidth}p{0.24\linewidth}p{0.22\linewidth}}
\toprule
\textbf{Asset} & \textbf{Role in this paper} & \textbf{Citation/source} & \textbf{License or terms} \\
\midrule
LoCoMo & Long-term conversational-memory benchmark for QA evaluation & \citet{maharana2024locomo} & CC BY-NC 4.0 release license \\
LongMemEval-S & Dynamic multi-session memory benchmark for QA evaluation & \citet{wu2024longmemeval} & MIT-licensed public release \\
DailyDialog & Conversation data used to train the discrete codebook component & \citet{li2017dailydialog} & CC BY-NC-SA 4.0 release license \\
T5-base & Encoder initialization for isomorphic semantic quantization & \citet{raffel2020t5} & Apache-2.0 release license \\
bge-base-en-v1.5 & Dense embedding backbone for vector retrieval & \citet{bge2024} & MIT public model-release terms \\
GPT-4o-mini API & Memory writing, answer generation, and LLM-as-a-judge evaluation & \citet{openai2024gpt4omini} & OpenAI service terms \\
Qwen3-30B-A3B-Instruct-2507 & Open-weight LLM backbone in the main evaluation & \citet{qwen3_2025} & Apache-2.0 model release \\
Mem0 & Baseline long-term memory system & \citet{mem0_2025} & Apache-2.0 public repository license \\
LangMem & Baseline long-term memory system & \citet{langmem2025} & MIT public repository license \\
A-MEM & Baseline agentic memory system & \citet{amem_2025} & MIT public repository license \\
MemoryOS & Baseline hierarchical memory system & \citet{memoryos_2025} & Apache-2.0 public repository license \\
LightMem & Baseline lightweight memory system & \citet{lightmem_2025} & MIT public repository license \\
\bottomrule
\end{tabular}%
}
\end{table}

\end{document}